\documentclass{article} % For LaTeX2e
\usepackage{iclr2025_conference,times}

\usepackage{amsmath,amsfonts,bm}

\def\eqref#1{equation~\ref{#1}}
\def\1{\bm{1}}

\DeclareMathAlphabet{\mathsfit}{\encodingdefault}{\sfdefault}{m}{sl}
\SetMathAlphabet{\mathsfit}{bold}{\encodingdefault}{\sfdefault}{bx}{n}

\usepackage{hyperref}
\usepackage{url}

\title{Mask in the Mirror: Implicit Sparsification}

\author{Tom Jacobs \\
CISPA Helmholtz Center for Information Security \\
\texttt{tom.jacobs@cispa.de}\\
\And
Rebekka Burkholz \\
CISPA Helmholtz Center for Information Security \\
\texttt{burkholz@cispa.de} \\
}

\iclrfinalcopy % Uncomment for camera-ready version, but NOT for submission.
\begin{document}

\maketitle

\begin{abstract}

Continuous sparsification strategies are among the most effective methods for reducing the inference costs and memory demands of large-scale neural networks. A key factor in their success is the implicit $L_1$ regularization induced by jointly learning both mask and weight variables, which has been shown experimentally to outperform explicit $L_1$ regularization. We provide a theoretical explanation for this observation by analyzing the learning dynamics, revealing that early continuous sparsification is governed by an implicit $L_2$ regularization that gradually transitions to an $L_1$ penalty over time. Leveraging this insight, we propose a method to dynamically control the strength of this implicit bias. Through an extension of the mirror flow framework, we establish convergence and optimality guarantees in the context of underdetermined linear regression. Our theoretical findings may be of independent interest, as we demonstrate how to enter the rich regime and show that the implicit bias can be controlled via a time-dependent Bregman potential. To validate these insights, we introduce PILoT, a continuous sparsification approach with novel initialization and dynamic regularization, which consistently outperforms baselines in standard experiments.
\end{abstract}

\section{Introduction}

Deep learning continues to impress across disciplines ranging from language and vision \citep{dalle2} to drug design \citep{drugdesign,alphafold} and even fast matrix multiplication \citep{matrix}. 
These accomplishments come at immense costs, as they rely on increasingly large neural network models.
Moreover, training such massive models with first-order methods like variants of Stochastic Gradient Descent (SGD) is a considerable challenge and often requires large-scale compute infrastructure \citep{kaack2022aligning}. Even higher costs are incurred at inference time, if the trained models are frequently evaluated \citep{wu2022sustainable,luccioni2023power}.

Sparsifying such neural network models is thus a pressing objective. 
It not only holds the promise to save computational resources, it can also improve generalization \citep{frankle2019lottery,paul2023unmasking}, interpretability \citep{chen2022can,hossain2024tickets}, denoising \citep{prune-regularize,wang2023searching}, and verifiability \citep{Narodytska2020In,albarghouthi2021introduction}. 
However, at its core is a hard large-scale nested optimization problem combining multiple objectives.
In addition to minimizing a typical neural network loss $\min_{w \in \mathbb{R}^n} f(w)$ (and its generalization error), we wish to rely on the smallest possible number of weights, effectively minimizing the $L_0$ norm $\min_{w \in \mathbb{R}^n} ||w||_{L_0}$.
This is an NP-hard problem that is also practically hard to solve due to its mixed discrete and continuous nature.
This becomes more apparent when we reformulate it like the best performing sparsification methods.

Among such approaches that achieve high sparsity while maintaining high generalization performance are continuous sparsification methods and iterative pruning strategies. 
%albeit being computationally expensive, are iterative pruning strategies and continuous sparsification methods
These methods explicitly identify for each weight parameter $w$ of a neural network a binary mask $m \in \{0,1\}$ that signifies whether a parameter is pruned. Thus, a parameter is set to zero ($m=0$) or not ($m=1$), effectively parameterizing the network with parameters $x=m \odot w$, where $\odot$ is the pointwise mutliplication (Hadamard product).  
The introduction of the additional mask parameters $m$ turns the sparsity objective into a discrete $L_1$ penalty of $m$ as $||w \odot m||_{L_0} = \sum_i m_i$ subject to $m_i \in \{0,1\}$, where $\odot$ denotes elementwise multiplication and we assume that $w \neq 0$ if $m=1$. 
The $L_1$ objective is already more amenable to continuous optimization than the original $L_0$ objective \citep{louizos2018learning}.
Nevertheless, the big challenge arises from the fact that $m$ is binary.

Continuous sparsification addresses this issue by relaxing the optimization problem to continuous or even differentiable variables $m$, often with $m \in {[0,1]}$ by learning a parameterization $m=g(s)$ with $g$: $\mathbb{R} \rightarrow {[0,1]}$, e.g., like a sigmoid. 
%We later discuss different conceptual choices of $g$ and their potential limitations for sparsification from the perspective of mirror flows. 
This way, the problem becomes solvable with standard first-order optimization methods.
Yet, moving from the continuous space back to the discrete space is error-prone. 
Regularizing and projecting $m$ towards binary values $\{0,1\}$ is generally problematic, requires careful tuning, and often entails robustness issues.

However, this projection step is not necessarily required in a parameterization $m \odot w$, where $m$ can freely attain values in $\mathbb{R}$.
\citet{ziyin2023spredsolvingl1penalty} show that a loss with this parameterization $m \odot w$ combined with weight decay $\alpha \left(||m||^2_{L_2} + ||w||^2_{L_2}\right)$ is equivalent to LASSO and thus an explicit $L_1$-regularization \citep{Ziyin2023SymmetryIS}.
Yet, their proposed method spred
%the method spred, which combines $m \odot w$ with an additional weight decay $\alpha \left(||m||^2_{L_2} + ||w||^2_{L_2}\right)$, which appears to be equivalent to LASSO. 
significantly outperforms LASSO, which suggests that the posed equivalence of optimized objectives cannot explain this success.

% Moreover, spred is shown to be competitive with STR \citep{kusupati20}, a state-of-the-art continuous sparsification method, which optimizes implicitly for the $L_1$-penalty. This involves a thresholding operation,
% %To utilize this insight for continuous sparsification,  
% which controls the strength of the implicit $L_1$-penalty.
% Due to the implicit nature of the $L_1$-penalty STR solves the hierarchical optimization problem:
As we show, an important, but overlooked, distinguishing factor is the fact that continuous sparsification with $m \odot w$ parameterization is driven by an implicit rather than an explicit regularization. 
This implies that in a sufficiently overparameterized setting, the following hierarchical optimization problem is solved: 
\begin{equation}\label{introduction : new opt}
    \min_{x \in \mathbb{R}^n : f(x) = 0}||x||_{L_1}.
\end{equation}
%in sufficiently overparameterized settings. 
The main idea follows the same philosophy as the lottery ticket hypothesis \citep{frankle2019lottery}.
From a range of models which all attain optimal training loss, it prefers the sparsest model. 
In other words, we aim to find a subnetwork with a similar accuracy as a dense network. 
%Accordingly, we call our approach PILoT.
Instead of having two competing objectives, we enforce cooperation between the two objectives by subjugating the sparsification. 
While this formulation can have advantages, if we can attain zero training loss ($f(x) = 0$), it might still be equivalent to an explicit regularization.
%$\min_{x \in \mathbb{R}^n : f(x) = 0}||x||_{L_1}$

%Recently, \citep{ziyin2023spredsolvingl1penalty} have proposed spred, which uses the paramterization $m \odot w$ with weight decay.
%Together with \citep{Ziyin2023SymmetryIS} they have shown that this is equivalent to LASSO. 
%This method removes the need for projecting back to the discrete space.
%Nevertheless, the equivalence to LASSO does not explain why it performs significantly better.
%Yet, how can we control the sparsity
%Our analysis implies that also other continuous sparsification approaches, including STR, ..., share similar implicit biases
The real potential of continuous sparsification and its induced implicit bias (in particular in non-convex settings) becomes apparent when we study the corresponding learning dynamics.
Our theoretical analysis of (stochastic) gradient flow applied to $m \odot w$ reveals that the dynamics differ fundamentally from the ones of LASSO, where redundant features are sparsified exponentially fast. 
%Our theoretical analysis of (stochastic) gradient flow applied to $m \odot w$  provides an explanation, as the dynamics differ fundamentally from the ones of LASSO, where redundant features are sparsified exponentially fast. 
Instead, we show by extending the mirror flow framework that a dynamic explicit weight decay can move the implicit bias from $L_2$ to $L_1$ during training.
In consequence, the sparsification becomes effective only relatively late during training, allowing the overparameterized model to first attain high generalization performance.
The dynamics resemble thus a successful strategy that is applied across continuous and iterative pruning methods, which all acknowledge and realize the premise that training overparameterized models 
before they are sparsified usually leads to significant performance benefits \citep{Frankle2018TheLT,paul2023unmasking,gadhikar2024masks}.

Our analysis extends the implicit bias framework, which covers different parameterizations \citep{pesme2021implicit, gunasekar2020characterizing, woodworth2020kernel, Li2022ImplicitBO} and the well-known fact that standard gradient flow has an implicit $L_2$ bias \citep{nemirovskiĭ1983problem, BECK2003167}. 
%The implicit bias has been used to study 
Another common use of the framework has been to study when training dynamics enter the so-called rich regime, which is responsible for improved feature learning. 
%Different parameterizations though induce a different implicit bias \citep{pesme2021implicit, gunasekar2020characterizing, woodworth2020kernel, Li2022ImplicitBO} in the (stochastic) gradient flow framework.
%\citep{pesme2021implicit, gunasekar2020characterizing, woodworth2020kernel, Li2022ImplicitBO} in the (stochastic) gradient flow framework.
%We theoretically extend the framework by proposing a 
In this context, we study two main innovations.
a) As we show, the explicit regularization (i.e. weight decay on $m$ and $w$) guides the strength of the implicit bias.
The fact that this makes the implicit bias tuneable makes it practically relevant for sparsification, as we have to be able to reach a target sparsity.
b) By proposing a dynamic regularization (rather than a common static one), we obtain control over the transition speed from $L_2$ to $L_1$ regularization, which is crucial for performance gains in the high sparsity regime and enables us to enter the rich regime.
%Our main proposal is to study the impact of explicit regularization on the implicit one
%make the explicit weight decay on $m$ and $w$ dynamic rather than fixed as common.
% By proposing a dynamic regularization on $m$ and $w$, we make the resulting implicit regularization tuneable.
% The fact that it is tuneable (and not fixed as common) makes it practically useful for sparsification, where we have to control the speed of .  
%we can obtain sparser models that prioritize our 
%This regularization leads to even sparser lottery tickets, while still prioritizing the main optimization goal: accuracy.
% In this sense, it can be interpreted as tuneable implicit regularization.
%It is defined by a time-dependent Bregman potential, which controls the bias and is potentially of independent theoretical interest.
%\textcolor{blue}{Thus, we} 
From a conceptual point of view, we unite explicit and implicit bias within a time-dependent Bregman potential, which is potentially of independent theoretical interest.

While our general derivations provide insights into various continuous sparsification approaches, including STR \citep{kusupati20}, spred \cite{ziyin2023spredsolvingl1penalty}, or \citep{savarese2021winning}, we also utilize them to propose a new improved algorithm, PILoT (Parametric Implicit Lottery Ticket). 
PILoT combines the $m \odot w$ parameterization with a dynamic regularization and an initialization that enables sign flips.
Such sign flips are key to effective sparse training \citep{gadhikar2024masks}, but are not feasible with the spred initialization.
%Furthermore, we show that spred is unable to switch signs which is key in sparsification \citep{gadhikar2024masks}. 
%Out new initialization allows for such sign flips.
%In addition, the dynamic regularization allows us to improve significantly over spred on multiple benchmarks.
%We show that explicit and implicit bias can be united with a time-dependent Bregman potential.
The dynamic regularization and thus implicit bias leads us to outperform state-of-the-art baselines in particular in the high-sparsity regime, as we demonstrate in extensive experiments. 

In summary, we make the following \textbf{contributions:} 
\begin{itemize}
    \item We gain novel insights into continuous sparsification by highlighting its implicit bias towards sparsity induced by doubling the number of trainable parameters. In particular, we explain the effectiveness of spred \citep{ziyin2023spredsolvingl1penalty}, which is based on $m \odot w$.
    %\item In doing so, we extend the mirror flow framework and derive a time-dependent Bregman potential.
    \item To the best of our knowledge, we are the first to introduce the implicit bias with an explicit regularization resulting in a mirror flow with a time-dependent Bregman potential.
     \item We provide convergence results for (quasi)-convex loss functions (Theorem \ref{thm : convergence text}) and optimality for underdetermined linear regression (Theorem \ref{exp reg : optimality time dep}) with time-dependent Bregman potential.
    \item Improving results by \citep{Alvarez_2004, Li2022ImplicitBO}, we replace convexity with the Polyak-\L ojasiewicz inequality, quasi-convexity and a growth condition on the Bregman potential (see Theorem \ref{Thm : Arora 4.14 improvement}).
    \item Using our extensions of the mirror flow framework, we propose a new continuous sparsification method, PILoT, which controls the implicit regularization dynamically moving from $L_2$ to $L_1$. Its initialization enables sign flips in contrast to spred. 
    \item In experiments for diagonal linear networks and vision benchmarks (including ImageNet), PILoT consistently outperforms baseline sparsification methods such as STR and spred, which demonstrates the utility of our theoretical insights.
\end{itemize}

\subsection{Related work}\label{section : related work}
\textbf{Neural network sparsification.}
A multitude of neural network sparsification methods have been proposed with different objectives \citep{handbook}. 
A popular objective is, for instance, to save computational and memory costs primarily at inference, or also during training, which is linked to the time of pruning, i.e., initially \citep{frankle2021review,snip,synflow,grasp,pham-paths,patil2021phew, synflow,random-pruning-vita,er-paper,plant}, early during training \citep{evci-rigl,sparse-momentum-dettmers}, during training like continuous sparsification \citep{rare-gems,kusupati20,savarese2021winning,peste2021acdc} or within multiple pruning-training iterations \citep{han-efficient-nns,Frankle2018TheLT,earlybird,rewindVsFinetune,gadhikar2024masks}.
Other distinguishing factors are which type of sparsity the methods seek, if they focus on saving computational resources and memory in specific resource-constrained environments or, which methodological approach they follow.

\textbf{Unstructured sparsity.}
In this work, we focus on unstructured sparsity, i.e., the fraction of pruned weights, and thus seek to remove as many weight entries as possible, which can achieve generally the highest sparsity ratios while maintaining high generalization performance. 
Structured sparsity, which usually obtains higher computational gains on modern GPUs \citep{kuzmin2019taxonomy,NIPS2016_41bfd20a,struct-dst}, could also be realized in the continuous sparsification setting, for instance, by learning neuron-, group, or even layer-wise masks.
Yet, this would not enjoy the same theoretical benefits as we derive here by showing that the unstructured continuous $m \odot w$ parameterization induces a mirror flow, whereas for example the neuron-wise mask does not (see Section \ref{section : neuronwise}). 

\textbf{Iterative pruning.}
Iterative pruning is often motivated by the Lottery Ticket Hypothesis (LTH) \citep{Frankle2018TheLT}, which conjectures the existence of sparse subnetworks of larger dense source networks that can achieve the same accuracy as the dense network when trained \citep{frankle2021review, liu2024survey, malach2020proving, orseau2020logarithmic, pensia2020optimal, uniExist, plant, convexist, depthexist, cnnexist, equivariantLT}.
In addition to the sparse structure, iterative pruning tries to identify a trainable parameter initialization, indirectly also implementing an approximate $L_0$-regularization.
In repeated prune-train iterations, trained weights are thresholded according to an importance score like magnitude.
Afterward, the remaining parameters are free to adapt to data in a new training run and not be regularized by a sparsity penalty (like $L_1$).
Our proposal PILoT can be combined with such iterative schemes.
The experiments show that it boosts performance of state-of-the-art schemes like Weight Rewinding (WR) \citep{DBLP:journals/corr/abs-1912-05671}, and Learning Rate Rewinding (LRR) \citep{maene2021understanding}.  

\textbf{Continuous sparsification.}
Continuous sparsification characterizes a collection of methods that can compete with iterative pruning techniques, while often requiring fewer training epochs \citep{rare-gems,  kusupati20}.
In one of the first proposals by \citep{savarese2021winning}, the mask is relaxed to a continuous variable. 
In general, continuous sparsification can be combined with a probabilistic approach where $m$ is interpreted as a probability 
\citep{louizos2018learning,zhou2021efficient,zhou2021effective}.  
Other parameterizations of $m$ that are not restricted to the range ${[0,1]}$ (e.g. Powerpropagation) can also be utilized to regularize towards higher sparsity \citep{schwarz2021powerpropagation}.
Yet, they are usually combined with projection to map $m$ to a binary mask.
The spred algorithm \citep{ziyin2023spredsolvingl1penalty} removes any projections and shows that $m \odot w$ with weight decay solve a LASSO objective. 
To explain its performance gain over LASSO, we extend 
the mirror flow framework and find an explanation in the training dynamics. 
%that implies a dynamic transition from an implicit $L_2$ to $L_1$ regularization.
%We derive why spred can outperform LASSO and improve upon the algorithm with PILoT, with the help of our novel framework. 
Our extension, PILoT, dynamically adjusts the weight decay and induces a transition from an implicit $L_2$ to $L_1$ regularization. 
This enables it to outperform
%Nevertheless, the current state of the art is still 
the state-of-the-art method STR \citep{kusupati20} in the high-sparsity regime.
For a survey of other methods see \citep{kuznedelev2023accurate}.

\textbf{Implicit bias.}
The implicit bias of (S)GD is a well-studied phenomenon \citep{pmlr-v125-chizat20a, Li2022ImplicitBO, woodworth2020kernel, gunasekar2020characterizing, gunasekar2017implicit, chou2024induce, vaškevičius2019implicitregularizationoptimalsparse, li2023implicitregularizationgroupsparsity, Zhao_2022, li2021implicitsparseregularizationimpact} and can in certain cases be described by a mirror flow or mirror descent (in the discrete case with finite learning rate) \citep{Li2022ImplicitBO}. 
Originally, mirror descent was proposed to generalize gradient descent and other first-order methods in convex optimization \citep{Alvarez_2004, Rockafellar1970ConvexA, Boyd2009ConvexO, nemirovskiĭ1983problem, BECK2003167}. 
Moreover, it has been used to study the implicit regularization of SGD in diagonal linear networks \citep{pesme2021implicit,Even2023SGDOD}.
More recently, it also has been applied to analyze the implicit bias of attention \citep{sheen2024implicit}. 
While \citep{Li2022ImplicitBO} has shown that different parameterizations have a corresponding mirror flow, we find that $m \odot w$ with our proposed explicit regularization, PILoT, gives rise to a corresponding time-dependent mirror flow. 
Its time dependence gives us means to control the implicit bias, while still achieving convergence.
%Time-dependent 
Time-dependent mirror descent has so far only been studied in the discrete case as a general possibility \citep{radhakrishnan2021linear}. 
The time dependence also naturally arises in SDE modelling, yet, without control of the implicit bias \citep{pesme2021implicit, Even2023SGDOD}. 
Here, we not only highlight a practical use case for time-dependent Bregman potentials, we also derive a way to control and exploit it.  

\textbf{Optimization and convergence proofs.}
Loss landscapes and the convergence of first-order methods is a large field of study \citep{10.1007/978-3-319-46128-1_50, fehrman2019convergence} in its own right. 
We draw on literature that shows convergence by using the Polyak-\L ojasiewicz inequality \citep{wojtowytsch2021stochastic, dereich2024convergence}, which is a more realistic assumption in the deep learning context than, for example convexity, because it can hold locally true for non-convex loss functions that are common in machine learning.

\section{Controlling the implicit bias with explicit regularization}\label{section: Theory}

\textbf{Structure of theoretical exposition.}
Our first goal is to advance the mirror flow framework from \citep{Li2022ImplicitBO} to incorporate time-dependent regularization. This is a key innovation that forms the foundation of our proposed PILoT algorithm (Algorithm \ref{alg:cap}). 
The dynamical description also covers constant regularization, as implemented in the spred algorithm.
%In this section we extend the mirror flow framework as in \citep{Li2022ImplicitBO} to time dependend mirror flow.
%This lays the groundwork for PILoT (Algorithm \ref{alg:cap}). 
% We begin by demonstrating how the mirror flow framework integrates with time-dependent regularization, in case of the parameterization $m \odot w$ combined with time-varying weight decay.
%We first explain how the mirror flow framework can be used to improve over (constant) explicit regularization.
%Furthermore, we show that for the parameterization $m \odot w$ with time depending weight decay corresponds to a time dependent Bregman potential.
%We begin by demonstrating how the mirror flow framework integrates with time-dependent regularization, in case of the parameterization $m \odot w$ combined with time-varying weight decay.
We begin by integrating time-dependent regularization in the mirror flow framework in the case of the parameterization $m \odot w$ combined with time-varying weight decay.
This corresponds to a time-dependent Bregman potential, enabling a more dynamic and powerful form of implicit regularization. 
The implicit regularization becomes controllable and moves from an $L_2$ to an $L_1$ regularization.
Building on this, Theorem \ref{thm : time dep breg} rigorously characterizes this process within this extended framework, offering new insights into the sparsification process.
% We identify two shortcomings of the mirror flow framework: a) The implicit bias is centered at initialization; and b) the scale of the initial parameters (with $x_0=0$) is required to be exponentially small to induce an $L1$-regularization.
% %\textcolor{blue}{The insights lead to solving the two shortcomings of the mirror flow framework.}
%Next, the extended framework allows us to characterize the changing implicit bias in Theorem \ref{thm : time dep breg}.
Then Theorem \ref{thm : convergence text} establishes convergence to a solution of the original optimization problem. 
Sparsity is still attained according to Theorem \ref{thm : time dep breg}, which can also be observed in the gradient flow in Eq.~(\ref{pilot : gf}) of PILoT.
For diagonal linear networks, which is an analytically tractable setting, we also prove optimality, as stated by Theorem \ref{exp reg : optimality time dep}.
%Furthermore, we show optimality for diagonal linear networks
This highlights a mechanism how our method PILoT improves over spred, since spred cannot reach optimality.

\textbf{Optimization problem.}
Consider the following time-dependent optimization problem for a loss function $f : \mathbb{R}^n \rightarrow \mathbb{R}$:
\begin{equation}\label{PILoT: opt problem}
    \min_{m, w \in \mathbb{R}^n} f(m \odot w) + \alpha_t \left(||m||_{L_2}^2 + ||w||_{L_2}^2\right).
\end{equation}
where $\alpha_t \geq 0$ can change during training.
In contrast, \citep{ziyin2023spredsolvingl1penalty} set $\alpha_t = \alpha$ constant and show that Eq.~(\ref{PILoT: opt problem}) is equivalent to the LASSO objective.
Why does spred tend to outperform LASSO then?
%As mentioned in the introduction, this does not explain why it performs better than LASSO. 

\textbf{Seeking answers in the training dynamics.}
The gradient flow associated with minimizing the continuously differentiable loss function $f$ is:   $d x_t = -\nabla f(x_t) dt$, $x_0 = x_{\text{init}}$.
Using this gradient flow framework, \citep{Li2022ImplicitBO} show that a reparameterization or overparameterization of the parameters $x$ leads to a mirror flow. 
A mirror flow informally minimizes a potential in the background, for example, the $L_1$ or $L_2$-norm. In contrast, explicit regularization forces a direct trade-off. 
The no need for a trade-off becomes clear in the convergence and optimality theorem.

\textbf{Mirror flow.}
Concretely, to define a mirror flow, let $R : \mathbb{R}^n \rightarrow \mathbb{R}$ be a differentiable function. It is described by 
\begin{equation*}\label{intro : mirrorflow}
    d\nabla R(x_t) = - \nabla f(x_t) dt, \qquad x_0 = x_{\text{init}}.
\end{equation*}
\citep{Li2022ImplicitBO} provide sufficient conditions for a paramerization $g$: $M \rightarrow \mathbb{R}^n$ to induce a mirror flow, where $M$ is a smooth sub-manifold in $\mathbb{R}^D$ for $D \geq n$. 
The parameterization $m \odot w$ falls into this category. 
% Moreover, the resulting potential is an interpolation between the $L_1$ and $L_2$ norm.
% Therefore, we want to use the parameterization to induce sparsity by steering the interpolation towards $L_1$.
% Nevertheless, the potential $R$ has its global minima at the initialization $x_0$, this holds true for reparameterizations \citep{Li2022ImplicitBO}.
The corresponding potential $R$ is either close to the $L_1$ or $L_2$ norm depending on the initialization of $m_0$ and $w_0$.
If we could steer it towards $L_1$, we could therefore induce an implicit regularization towards sparsity, yet, not without issues.

\textbf{Two caveats and their solution.}
a) The potential $R$ attains its global minimum at the initial $x_0$ (and not 0). This also holds for other reparameterizations \citep{Li2022ImplicitBO}. In consequence, we would not promote actual sparsity for $x_0 \neq 0$.
b) To induce $L_1$ regularization and enter the rich regime, the initialization of both $m_0$ and $w_0$ would need to be exponentially small \citep{woodworth2020kernel}. 
%This exponentially small initialization corresponds to the rich regime.
% \begin{itemize}
%     \item The potential $R$ attains its global minimum at the initialization $x_0$ (and not 0). This also holds for other reparameterizations \citep{Li2022ImplicitBO}. In consequence, we would not promote actual sparsity for $x_0 \neq 0$.
%     \item To induce the $L_1$ regularization, the initialization of both $m_0$ and $w_0$ would need to be exponentially small \citep{woodworth2020kernel}. 
% \end{itemize}

%Firstly, the potential $R$ attains its global minimum at the initialization $x_0$ (and not 0)\textcolor{blue}{. This} also holds for other reparameterizations \citep{Li2022ImplicitBO}.
%In consequence, we would not promote actual sparsity.

%Secondly, to induce the $L_1$ regularization, the initialization of both $m_0$ and $w_0$ would need to be exponentially small \citep{woodworth2020kernel}. 

The explicit dynamic regularization of PILoT in Eq.~(\ref{PILoT: opt problem}) solves both of these problems, as we show next.
The corresponding mirror function $R$ is stated in Theorem \ref{thm : implicit bias} and the corresponding convergence and optimality theorems in Theorem \ref{Thm : Arora 4.14} and \ref{exp reg : optimality}. 
%These results motivate the use and extension of the mirror flow framework.

\textbf{Dynamic regularization.}
Next we present the main result, the dynamical description with time-dependent regularization.
%Our goal is to control the implicit bias, i.e., transition from an $L_2$ to an $L_1$ bias.
%This can be established by controlling the regularization strength $\alpha_t$.
The exact dynamics are described by the time-dependent mirror flow and is derived in Theorem \ref{thm : time dep breg}.
%\begin{itemize}
 %   \item explain mirror flow and time dep mirror flow
 %   \item main theorems from section 3
 %   \item Put results section 2 in appendix
 %   \item merge section 2 and 3 of previous submission
 %   \item link to convergence of algorithm
%\end{itemize}

%\begin{itemize}
%    \item explain mirror flow and its downsides
%    \item refer to appendix for more information
    
%\end{itemize}
\begin{theorem}\label{thm : time dep breg}
Let $|w_{0,i}| < m_{0,i}$ for all $i \in [n]$, the time-dependent Bregman potential is given by 
    \begin{equation}\label{expreg : hypen}
	R_{a_t} (x) = \frac{1}{2}\sum_{i = 1}^n x_i \text{arcsinh}\left( \frac{x_i}{ a_{t,i}} \right) - \sqrt{x_i^2 + a_{t,i}^2} - x_i \text{log} \left(\frac{u_{0, i}}{v_{0, i}}\right),
\end{equation}
with $a_{t,i} = 2u_{0,i} v_{0,i} \text{exp}\left(- 2\int_0^t \alpha_s ds\right)$ and $u_{0,i} = \frac{m_{0,i} + w_{0,i}}{\sqrt{2}}$ and $v_{0,i} = \frac{m_{0,i} -w_{0,i}}{\sqrt{2}}$. 
The gradient flow of $x_t = m_t \odot w_t$ induced by Eq.~(\ref{PILoT: opt problem}) then satisfies
\begin{equation*}\label{exp reg : Mirror flow}
	d \nabla R_{a_t} (x_t) = -\nabla f(x_t) dt, \qquad x_0 = m_0 \odot w_0.
\end{equation*}
\end{theorem}
%\begin{proof}
\textbf{Proof.} 
The proof is given in the appendix. The main steps are:
a) Deriving the evolution of the gradient flow (Lemma \ref{appendix : lemma flow}).
b) Showing that it satisfies the time-dependent mirror flow (Lemma \ref{appendix : lemma mirror}). Note that step a) also derives Eq.~(\ref{pilot : gf}).
%\end{proof}
% \begin{itemize}
%     \item Deriving the evolution of the resulting gradient flow (Lemma \ref{appendix : lemma flow}).
%     \item Showing it satisfies the time-dependent mirror flow (Lemma \ref{appendix : lemma mirror}).
% \end{itemize}

\begin{wrapfigure}[17]{R}{0.35\textwidth}
         \centering
         \includegraphics[width=0.35\textwidth]{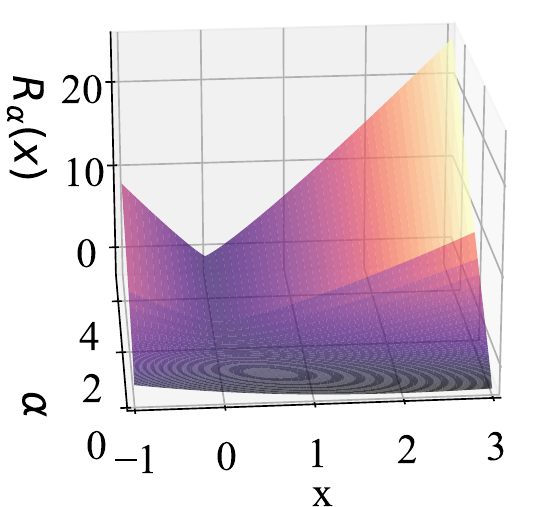}
        \caption{Evolution of the time-dependent Bregman potential. $\alpha = \int_0^t \alpha_s ds$ is the exponent of $a_t$.}
       \label{fig: Breg evol}
\end{wrapfigure}

Observe that the potential in Eq.~(\ref{expreg : hypen}) now depends on $a_t$. This changes the global minimum to
%This has the following effect on the position of the global minimum:
\begin{equation*}
    \nabla R_{a_t}(x) = 0 \Leftrightarrow x = \text{exp}\left(- 2\int_0^t \alpha_s ds\right) \odot m_0 \odot w_0 .
\end{equation*}
Thus, we gain control over the positional implicit bias, solving our problem with the nonzero global minimum. 
Next, we characterize the asymptotic behavior, which we control in practice with $\alpha_t$, which determines also $a_t$. 
The asymptotics follows from Theorem 2 in \citep{woodworth2020kernel}.
%, solving one of our problems. 
%Furthermore, the asymptotic behavior is now controllable. 
For $a \rightarrow 0$ and $|\frac{x}{a}| \rightarrow \infty$, we receive
%$R_a(x) \sim \text{log}\left(\frac{1}{a}\right) ||x||_{L_1}$. 
\begin{equation*}
    R_a(x) \sim \text{log}\left(\frac{1}{a}\right) ||x||_{L_1}.
\end{equation*}
Interestingly, the term $x_i \text{log}\left(\frac{u_{0,i}}{v_{0,i}}\right)$ does not play a role in the asymptotics.
The reason is that $ \text{log}(\frac{1}{a})$ in front of the other term dominates. 
Figure \ref{fig: Breg evol} illustrates the asymptotics for the one dimensional case. 
We observe that our two previously identified mirror flow caveats can be resolved:
a) Increasing $a$ moves the minimum to the origin, leading to an $L_1$ regularization.
This implies initializing at zero with an exponentially small scaling (i.e. $a \rightarrow 0$) is not necessary.
b) Moreover, the regularization $\alpha_t$ has an exponential and time-dependent effect on $a_t$. 
It thus enables steering the dynamics towards an $L_1$ regularization at the desired speed. 
%In conclusion, our extension and novel analysis have discovered a promising initialization and revealed how explicit regularization solves the problems of the standard mirror flow framework.
In conclusion, our extension and novel analysis have revealed how explicit regularization solves the problems of the standard mirror flow framework.

\begin{remark}\label{exp reg : remark reg}
 At the end of LASSO training, the regularization can be turned off to enable the search for a better solution. 
 However, this risks losing the benefits of the regularization, for example, if the basin of attraction contains non-sparse critical points. 
 In contrast, for the $m \odot w$ reparameterization, the time-dependent Bregman potential steers (but does not force) the bias towards sparsity.
%Therefore, dynamically updating the regularization strength makes sense with the reparameterization. 
\end{remark}

\textbf{Convergence and optimality.}
It remains to be shown that convergence and optimality results transfer from the mirror flow framework to the time-dependent mirror flow framework.
For quasi-convex loss functions, we prove convergence to a critical point. 
For convex or quasi-convex functions that satisfy the PL-inequality, we derive convergence to a minimizer.
%bridge

\begin{theorem}\label{thm : convergence text}
    Assume $f$ is quasi-convex, $\nabla f$ is locally Lipschitz and $\text{argmin} \{ f(x) | x \in \mathbb{R}^n\}$ is non-empty. Assume $\alpha_t \geq 0$ for all $t \geq 0$ and that $\int_0^{t} \alpha_s ds < \infty$ for $t \in [0, \infty) \cup \{\infty\}$. Then as $t \rightarrow \infty$, $x_t$ converges to some critical point $x^*$.
    Furthermore, if $f$ is either convex or both quasi-convex and satisfies the PL-inequality in Eq.~(\ref{PL inequality}). 
    Then $x_t$ convergences to an interpolator $x^*$ that is a minimizer of $f$. 
    Furthermore, in the PL-inequality case, the loss converges linearly such that there is a constant $C > 0$ such that 
    \begin{equation}\label{exp reg : conv rate}
        f(x_t) -f(x^*)\leq B \text{exp}\left(-\lambda A_{\infty} t\right),
    \end{equation}
    where $B = \left(f(x_0) -f(x^*)\right)\text{exp}\left(C||x^*||_{L_2}\int_0^{\infty} \alpha_s ds\right)$ with $C$ depending on the smoothness of the loss function and $A_{\infty} = \min_i \left(m_{0,i}^2 - w_{0,i}^2\right) \exp \left(-2\int_0^{\infty} \alpha_s ds\right)$.
\end{theorem}
\textbf{Proof.} 
% The main steps are: a) Showing that the iterates are bounded and convergence to a critical point (Lemma \ref{appendix : itter bound}). b) Convergence of the loss (Theorem \ref{thm : convergence}). Notable tools are a time-dependent Bregman divergence used to bound the iterates and $\alpha_t \geq 0$ converges to zero, resulting in eventually a non-increasing  
%  evolution of the loss.
The main steps of the proof are to show that a) the iterates are bounded 
and converge to a critical point (Lemma \ref{appendix : itter bound}); b) the loss converges (Theorem \ref{thm : convergence}). A noteworthy tool is a time-dependent Bregman divergence, which we use to bound the iterates. Furthermore, we utilize that $\alpha_t \geq 0$ converges to zero, resulting eventually in a non-increasing evolution of the loss.

\textbf{Potential drawbacks.}
Theorem \ref{thm : convergence text} guarantees convergence in the case of implicit regularization.
Explicit $L_1$ regularization or spred, on the other hand, cannot achieve the same result due to constant regularization, which we will also highlight in experiments.
Regardless, note that the constant $B$ in Eq.~(\ref{exp reg : conv rate}) could be large.
Furthermore, to reach the implicit $L_1$-regularization, $a_{t}= \left( m_0^2 - w_0^2\right) \exp\left( -2\int_0^{t} \alpha_s ds \right)$ needs to be exponentially small similarly as in Theorem 2 in \citep{woodworth2020kernel}. 
These two potential drawbacks also reveal where the method will work, namely, in overparameterized settings where the solution $x^*$ should have less active parameters. 
Then $B$ is potentially relatively small.

\begin{remark}\label{remark : speed}
    If $\nabla f$ is one-sided inversely Lipschitz, a speed-up is possible. The quantity that needs to be bounded for convergence is $-\nabla f(x_t)^{\top} x_t$. In this case, we get
    \begin{equation*}
        -\nabla f(x_t)^{\top} x_t \leq -\nabla f(x_t)^{\top} x^* - ||x_t-x^*||^2_{L_2} \leq C ||x^*||_{L_2} - ||x_t-x^*||^2_{L_2} ,
    \end{equation*}
    where $C$ is the bound on the smoothness of the loss function $f$. This implies that when the interpolator $x^* \approx 0 $ is small, the right-hand side is negative, leading to a speed-up. This condition is also known as coercive.
\end{remark}

\textbf{Optimality.}
The main assumption in Theorem \ref{thm : convergence text} is $\alpha_t \rightarrow 0$, which ensures convergence to a minimizer of the original problem, while retaining sparsity depending on $a_{t} \rightarrow a_{\infty}$.
In the case of diagonal linear networks, we can even prove optimality with respect to the final Bregman potential $R_{a_{\infty}}$.
\begin{theorem}\label{exp reg : optimality time dep}
    In case of under-determined regression consider the loss function $f(x) = \Tilde{f}(Zx - Y)$. Assume $f$ satisfies the conditions with at least one of the convergence criteria of Theorem \ref{thm : convergence text}. Then $x_t$ converges to $x^*$ such that
    \begin{equation}\label{exp reg : opt L1 time dep}
        x^* = \text{argmin}_{Zx = Y} R_{a_{\infty}}(x).
    \end{equation}
\end{theorem}
\textbf{Proof.} We show that the KKT conditions of Problem (\ref{exp reg : opt L1 time dep}) are satisfied (Theorem \ref{appendix : optimality}).

% takeaway
%In this section we have extended the mirror flow framework.
%In addition, we have shown that convergence and optimality results transfer to the new setting.
%These insights are used in the next section to design the sparsification method PILoT (Algorithm \ref{alg:cap}).
\textbf{Take away.}
We have shown that our extended mirror flow framework can attain convergence and optimality in an analytically tractable scenario. Furthermore, from a practical side, it also gives us new tools to derive more promising continuous sparsification techniques with implicit regularization.
In the following section, we propose a way to dynamically control the transition from implicit $L_2$ regularization to $L_1$ during training with the help of the derived time-dependent Bregman potential.

\section{The algorithm: PILoT}
%In this section we present the proposed algorithm PILoT (Algorithm \ref{alg:cap}).
Like spred, our new algorithm PILoT (Algorithm \ref{alg:cap}) utilizes the parameterization $m \odot w$, but proposes a novel initialization and dynamic regularization schedule to control the transition from implicit $L_2$ to $L_1$ regularization.
%improve upon other continuous sparsification methods.
To attain the desired results in the original parameterization $x$, we first derive its gradient flow. 
%Theorem \ref{thm : convergence text}, the gradient flow informs us how to design the algorithm.

%To understand and improve upon this we derive the gradient flow in the original variable $x_t$.
%In addition, for our generalization we allow the regularization to be time dependent.
\textbf{Gradient flow.}
Essentially, the gradient flow follows from the analysis in Section \ref{section: Theory}.
%Inspired by Theorem \ref{thm : convergence text}, we design our algorithm to achieve better performance while guaranteeing convergence.
According to Theorem \ref{thm : convergence text}, we guarantee convergence by ensuring $\alpha_t \rightarrow 0$.
Concretely, the gradient flow for $x_t = m_t \odot w_t$ induced by Eq.~(\ref{PILoT: opt problem}) is given by:
\begin{equation}\label{pilot : gf}
    d x_t = -\sqrt{x_t^2 + a_t^2} \odot\left( \nabla f(x_t) + 2\alpha_t \frac{x_t}{\sqrt{x_t^2 + a_t^2}} \right) dt, \qquad x_0 = x_{init},
\end{equation}
where $a_t = \left( m_0^2 - w_0^2 \right) \exp\left(-2\int_0^t \alpha_s ds \right)$.  $m_0$ and $w_0$ have to be initialized such that $m_0 \odot w_0 = x_{init}$. 
Note that all operations are point-wise.
The derivation is based on the time-dependent mirror flow in Section \ref{section: Theory}.

\begin{remark}
    The gradient flow in Eq.~(\ref{pilot : gf}) allows us to make a direct comparison to the continuous sparsification method STR \citep{kusupati20}. 
    Instead of the soft thresholding operator, we have $\sqrt{x_t^2 + a_t^2}$. 
    The main difference is that STR does not change the magnitude of the gradient update outside of the (learnable) threshold, while both PILoT and spred actively change the magnitude depending on the magnitude of the weight.
    This active sparsification explains why spred and also PILoT can perform better in the high-sparsity regime.
\end{remark}
%On the hand of the gradient flow in (\ref{pilot : gf}) we can interpret why spred is better than LASSO and highlight where it can be improved.
\textbf{Spred.} 
The gradient flow in Eq.~(\ref{pilot : gf}) explains why spred performs better than LASSO and highlights where spred can further be improved.
Note that the balanced initialization of spred is defined such that $m_0^2 - w_0^2 = 0$ and the regularization is constant $\alpha_t = \alpha$.
Plugging this into Eq.~(\ref{pilot : gf}) gives 
\begin{equation}\label{pilot : gf spred}
    d x_t = -\sqrt{x_t^2} \odot\left( \nabla f(x_t) + 2 \alpha \text{sign}\left(x_t\right) \right) dt, \qquad x_0 = x_{init}.
\end{equation}
Compare this with the gradient flow of LASSO with regularization strength $2 \alpha$:
\begin{equation*}
      d x_t = -\left(\nabla f(x_t) + 2 \alpha \text{sign}\left(x_t\right) \right) dt, \qquad x_0 = x_{init}.
\end{equation*}
We observe that the main difference to the gradient flow of LASSO is the factor $\sqrt{x_t^2}$. 
This implies the considerable drawback that spred gradient flows cannot sign flip.
Therefore, it cannot reach the optimal solution or specific minimizers potentially.
Another way to see this is studying the evolution $x_t = x_0 \exp\left(-4\text{sign}\left(x_0\right) \int_0^t \nabla f(x_s) ds - 2\alpha t \right)$ satisfying Eq.~(\ref{pilot : gf spred}).
In practice, the absence of sign flips might be remedied by using a large learning rate and noise.
The evolution also explains why it can perform better than LASSO, as it decays redundant parameters exponentially faster instead of linearly. In other words, the gradient update is proportional to the magnitude of the parameter.
Therefore, the evolution of spred (Eq.~(\ref{pilot : gf spred})) can converge faster and come closer to zero than LASSO.

%\begin{remark}\label{pilot: param}
%    Our gradient flow analysis also leads to option to remove the overparameterization. We could discretize (\ref{pilot : gf} ) to get the update in the original parameters. 
%    Thus, removing the need for the parameters $m$ and $w$.
%\end{remark}

\textbf{PILoT.}
%Given our insights into the spred algorithm, we want to remedy these 
Our main goal is to remedy the caveats of the spred by inducing  the more general gradient flow in Eq.~(\ref{pilot : gf}).
The first improvement is to enable sign flips by changing the initialization to $m_0^2 - w_0^2 = \beta >0$, where $\beta$ denotes the scaling constant.
After discretizing Eq.~(\ref{pilot : gf}), the effective learning rate at initialization $x_0 = 0$ is $\eta |\beta|$, where $\eta >0$ is the learning rate.
Therefore, we use $\beta = 1$ in the experiments so that the learning rate $\eta$ is not altered.
%\textcolor{blue}{, leaving the learning rate as the single hyperparameter}. 
%\textcolor{blue}{
%$The choice is} %motivated by the discretization of the gradient flow.
%\textcolor{blue}{Therefore, the effective learning rate is $\eta$, leaving the learning rate as the single tunable hyperparameter}.

Our second and main improvement is induced by the time dependence of $\alpha_t$.
$\alpha_t$ controls the strength of both the implicit and explicit regularization via $a_t$. 
If $a_t >> x_t$, then the regularization term in Eq.~(\ref{pilot : gf}) resembles an $L_2$ instead of an $L_1$ norm.
Therefore decreasing $a_t$ moves the implicit regularization from $L_2$ to $L_1$.
Accordingly, we sparsify only mildly in early training epochs in contrast to spred.
%we sparsify gradually instead of abruptly at initialization.
We have shown this formally in Section \ref{section: Theory}.
Furthermore, convergence is covered by Theorem \ref{thm : convergence text} when $\alpha_t \rightarrow 0$. 
%The effect of the regularization remains during training, which is captured by the term $\sqrt{x_t^2 + a_t^2}$ in Eq.~(\ref{pilot : gf}).
%In consequence, PILoT leads to better accuracy while still having a lasting sparsification effect on the dynamics.
% Furthermore, where we show convergence in Theorem \ref{thm : convergence text} when $\alpha_t \rightarrow 0$ and that the effect of the regularization remains, which is captured by the term $\sqrt{x_t^2 + a_t^2}$ in (\ref{pilot : gf}).
% Thus leading to better accuracy while still having a lasting 
%As PILoT sparsifies less aggressively in early training rounds it can attain , while still converging relatively quickly, as we demonstrate in experiments.
%leads to better accuracy while still having a lasting sparsification effect on the dynamics.
Even when PILoT attains a similar sparsity as spred at the end of the training dynamics, it can usually still achieve a higher accuracy due to its improved training dynamics.

\textbf{Details on PILoT.}
The described design choices define Algorithm \ref{alg:cap}. 
%We use the new initialization to allow for sign flips.
%Furthermore, we dynamically update the regularization strength.
Our update of the regularization strength $\alpha_k$ depends on three quantities: a) the sparsity threshold $K$ for the weights (a hyperparameter), b) the training accuracy, and c) $\delta \geq 1$, the multiplicative factor to gradually increase or decrease the regularization strength.
%The update itself depends on $\delta \geq 1$.
The regularization strength (and thus sparsity) grows if the sparsity threshold has not been reached yet and the training accuracy has increased in the previous gradient update step.
%Moreover, the update is proportional to $\alpha_k$.
As the strength is adaptive, the algorithm is less sensitive to the initial strength $\alpha_0$. 
Note that the setting $\delta = 1$ and $\beta = 0$ corresponds to spred. 
Therefore, PILoT is a strict generalization of spred. 
In contrast to spred, however, after half of the training epochs, we decay the regularization strength regardless of whether the sparsity threshold $K$ is reached.
This guarantees convergence of the corresponding gradient flow, in accordance with Theorem \ref{thm : convergence text}.

%\begin{itemize}
 %   \item intuition, it follows from the gf that the influence of the regularization has a lasting sparsifying effect. Therefore, turning off the regularization still leads to a sparse solution in contrast with standard LASSO.
 %  \item The moving from an $L_2$ to $L_1$ implicit bias eases in the sparsification.
    
%\end{itemize}

\begin{algorithm}[ht]
\caption{PILoT}\label{alg:cap}
\begin{algorithmic}
\Require epochs $T$, schedule $\alpha_{init}$, initialization $x_{init}$, scaling constant $\beta$
\State Initialize $m_0, w_0$ such that $m_0 \odot w_0 = x_{init}$, $m_0^2 - w_0^2 = \beta$, $\delta\geq1$ and, $K$ %2u_0 \odot v_0
\State $\alpha_0 \leftarrow \alpha_{init}$
\State $Current\_training\_acc \leftarrow 0$
\State Set $\tilde{f}(m,w, \alpha_0) := f(m \odot w) + \alpha_0 \left(||m||^2_{L_2} + ||w||^2_{L_2}\right)$
\For{$k$ in $1 \hdots T$}
    \State $\left(m_k, w_k\right) =$ OptimizerStep$\left(\tilde{f}(m_{k-1},w_{k-1},\alpha_{k-1})\right)$ 
    \If{$Training\_acc \geq Current\_training\_acc$ and $||m_k \odot w_k||_{L_1} \geq K$ and $k \leq \frac{T}{2}$}
         \State $\alpha_{k} \leftarrow \alpha_{k-1} \delta$
    \Else 
         \State $\alpha_{k} \leftarrow \alpha_{k-1} / \delta$
    \EndIf
    \State $Current\_training\_acc \leftarrow Training\_acc$
\EndFor\\
\Return Model $f(x_T)$ with $x_T = m_T \odot w_T$
\end{algorithmic}
\end{algorithm}

\section{Experiments}
We demonstrate the effectiveness of PILoT in extensive experiments covering three different scenarios.
Firstly, we confirm our theoretical results on the gradient flow in Theorem \ref{exp reg : optimality time dep}.
Secondly, we compare PILoT with other state-of-the-art continuous sparsification methods such as STR \citep{kusupati20} and spred \citep{ziyin2023spredsolvingl1penalty} in a one-shot setting.  
In this context, we also isolate the individual contribution of our initialization. 
Finally, we combine PILoT with iterative pruning methods such as WR \citep{frankle2019lottery} and LRR \citep{maene2021understanding}. 
%\textcolor{blue}{We note that doubling the parameters may be concerning. Nevertheless, 

\textbf{Memory requirements.} 
As most other continuous sparsification approaches, note that PILoT doubles the number of parameters during training. Yet, according to \cite{ziyin2023spredsolvingl1penalty}, the training time of a ResNet50 with $m \odot w$ parameterization on ImageNet increases roughly by 5\% only and the memory cost is negligible if the batch size is larger than $50$. At inference, we would return to the original representation $x$ and therefore benefit from the improved sparsification.

%Note that even though PILoT doubles the number of parameters during training as most other continuous sparsification approaches, according to
%\cite{ziyin2023spredsolvingl1penalty}, the training time of a ResNet50 with $m \odot w$ parameterization on ImageNet increases roughly by $5\%$ only and the memory cost is negligible if the batch size is larger than $50$.
%Furthermore, at inference, we would return to the original but sparse representation $x$.
%Finally, we use PILoT to compress large neural networks.

%In this section, 

\begin{wrapfigure}{r}{0.45\textwidth}
         \centering
         \includegraphics[width=0.45\textwidth]{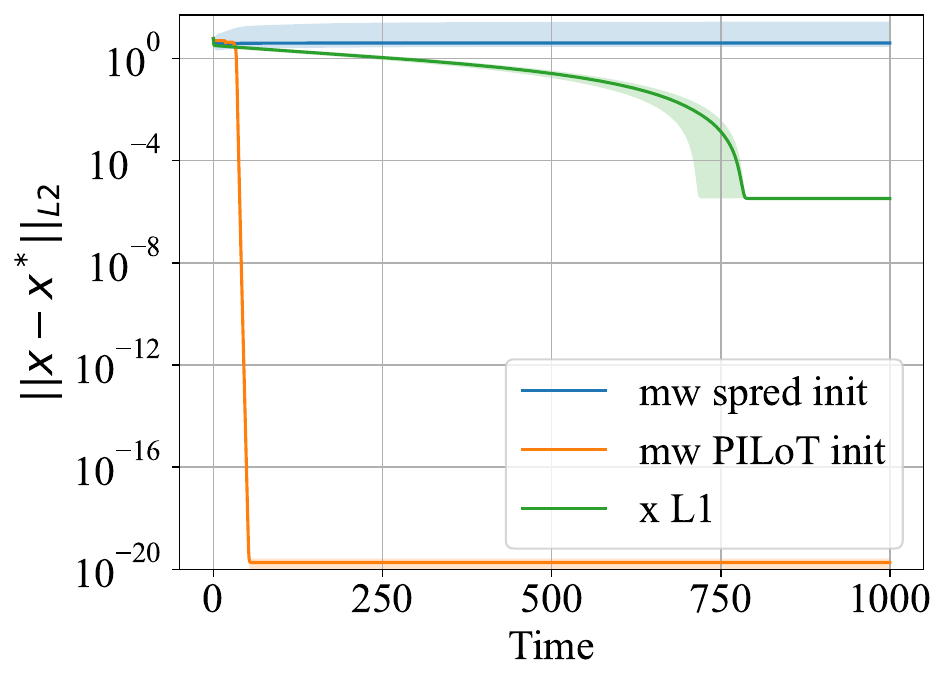}
        \caption{A simulation of gradient flow on a diagonal linear network is given for the different regularizations.}
        \label{fig: DLN reg}
\end{wrapfigure}

\paragraph{Diagonal Linear Network.}\label{section : DLN}
We have proven optimality for the analytically tractable setting of diagonal linear networks. 
Now we illustrate the benefit of our initialization and dynamic explicit regularization. 
Furthermore, we highlight the impact of a good dynamic schedule of the regularization strength $\alpha_t$. 
%The loss is given by 
%\begin{equation*}
%    f(x) = \frac{1}{2d}\sum_{j = 1}^d ( z_j^{\top} x - y_j)^2.
%\end{equation*}
We use $d = 40$ amount of data points with feature dimension $n = 100$ and sample $z_j \sim N(0, \mathbb{I}_n)$ for $j \in [d]$. 
The ground truth $x^*$ is set such that $||x^*||_{L_0} = 5$. 
Furthermore, the network parameters are initialized with $x_0 \sim N(0, \mathbb{I}_n \frac{1}{\sqrt{n}})$. 
The step-size is $\eta = 10^{-4}$ and the trajectories are averaged over 5 initializations. 0.95 confidence regions are indicated by shades.
%We show the distance between the ground truth and parameter value, i.e., $||x_t - x^*||_{L_2}$.
% compare with two others. 
%Figure \ref{fig: DLN init} highlights the importance of the initialization. 
%We see that with spred initalization $2u_0 v_0 = 0$, it is impossible to get close to the ground truth, for all regularization strengths.
%This highlights the necessity of our proposed PILoT initialization that allows us to provably reach the ground truth, even from a non-zero initialization $x_0$.
%Figure \ref{fig: DLN reg} shows the best regularization schedule for each initialization and LASSO. 
The mean squared error is used as loss function. 
%We report the distance of the evolution to the ground truth $||x_t - x^*||_{L_2}$ in Figure~\ref{fig: DLN reg}, which confirms our theoretical insights.
We report the distance to the ground truth $||x_t - x^*||_{L_2}$ over training time in Figure~\ref{fig: DLN reg}.
Two different initializations, i.e., the one of spred and of PILoT, and different regularization schedules are considered.
The schedules are described in Appendix \ref{dln appenix}.
Results for the best ones are shown in the figure and confirm our theoretical insights.
%Figure \ref{fig: DLN reg} reports results for 
%The best regularization schedule for each initialization and LASSO. 
%The initialization of $m$ and $w$ is critical for performance, as the inability to sign flip prevents spred from reaching the ground truth for all considered schedules. 
%Therefore, in the case of gradient flow, the balanced initialization should be avoided.
The inability to sign flip prevents spred from reaching the ground truth for all considered schedules. 
Furthermore, with a dynamic regularization, our PILoT initialization outperforms both spred and LASSO, reaching the ground truth. 
The best-performing schedule for the PILoT initialization is a geometrically decaying schedule, as also implemented in Algorithm \ref{alg:cap}.
%This serves as additional motivation for the regularization update step in Algorithm \ref{alg:cap}.
In contrast, for the other two methods, a constant regularization works best. 
This experimentally confirms Theorem \ref{exp reg : optimality time dep} and Remark \ref{remark : speed}.
Note that LASSO with gradient descent performs as expected.
%A constant schedule leads to the best performance, with a too small or large constant leading to worse performance.
%In addition, decaying schedules perform also worse supporting Remark \ref{exp reg : remark reg}.
Decaying schedules perform worse than constant ones supporting Remark \ref{exp reg : remark reg}.
%We note that even with a constant schedule our regularization is competitive with the other two settings, this is agreement with the analysis \citep{ziyin2023spredsolvingl1penalty, Ziyin2023SymmetryIS}. 
%In contrast, the reverse is not true.

% \textcolor{blue}{In conclusion,} the advantage of PILoT is \textcolor{blue}{the} access to the full parameter space, \textcolor{blue}{making sign flips possible}\textcolor{blue}{. Moreover,} the explicit regularization enables entering the rich regime\textcolor{blue}{. Thus,} obtaining an implicit $L_1$ regularization, as \textcolor{blue}{stated in Theorem \ref{thm : time dep breg} and} illustrated by 
% Figure \ref{fig: Breg evol}.

%shows the transition from an implicit $L_2$ regularization centered at the initialization towards an $L_1$ regularization.
%This illustrates Theorem \ref{thm : time dep breg}.

%\begin{itemize}
 %   \item 2 plots, 1 of evolution bregman pot and one of the gradient flow
 %   \item highlight spreds cannot signflip
 %   \item better acc than l1 for PILoT
 %   \item evolution highlight
%\end{itemize}

\paragraph{One-shot sparsification.}
%In this section, we present multiple one-shot experiments. 
%We first present experiments 
Firstly, we compare our method PILoT with STR, spred, and LASSO on CIFAR10 and CIFAR100 training a ResNet-20 or ResNet-18, respectively.
Furthermore, in the case of CIFAR10, we also implement the novel initialization ($m_0^2 -w_0^2 =1$) without dynamic regularization to isolate its benefits.
We consider two learning rates $\{0.1, 0.2\}$ and the weight decay range $\{1e-5, \hdots 1e-2\}$ for CIFAR10 and range $\{1e-4, \hdots 1e-3\}$ for CIFAR100 and always show the best result.
The same range for the regularization strength is explored for LASSO.
The other hyper-parameters are reported in the appendix.
Secondly, we train ResNet-50 on ImageNet with the setup of STR \citep{kusupati20} and compare directly with their results.
Furthermore, we implement both PILoT and spred in this setting.

%We first focus on the initalization where we use constant regularization i.e. $\alpha_t = \alpha$.
%The range of considered values for $\alpha$ is $\{1e-5, \hdots, 1e-2\}$.
%This experiment is done on CIFAR 10 with a ResNet20.

\begin{figure}[ht]
    \centering
    \includegraphics[width=0.95\textwidth]{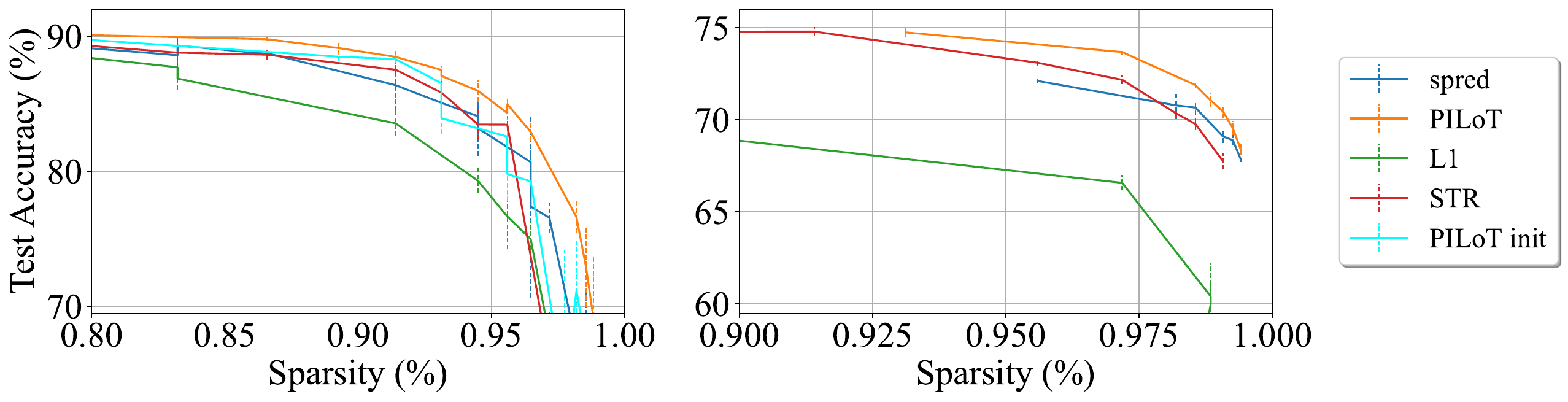}
    \caption{One-shot sparsification. Acc. versus sparsity for CIFAR10 (left) and CIFAR100 (right).}
    \label{fig: OS}
\end{figure}

%We implement PILoT in multiple settings. 
%We consider ResNet20 on CIFAR10 and ResNet50 on ImageNet.
Figure \ref{fig: OS} presents results for CIFAR10 and CIFAR100.
PILoT outperforms all other methods and is particularly effective in the high-sparsity regime.
Our PILoT initialization leads to improvements over spred and STR for medium levels of sparsity.
This supports our theoretical insight into the role of initializations and how they influence the implicit bias. 
In addition, STR is outperformed by spred in the high-sparsity regime, confirming the findings of \citep{ziyin2023spredsolvingl1penalty}.

%\begin{figure}[ht]
 %    \centering
 %    \
 %        \centering
 %        \includegraphics[width=\textwidth]{Figures/TrajectoryL1alpha.pdf}
 %        \caption{Example trajectory of $\alpha_k$ and $L_1$ norm.}
 %        \label{fig: L1 alpha trajectory}
 %    \end{subfigure}
 %    \hfill
 %    \begin{subfigure}[b]{0.4\textwidth}
 %        \centering
 %        \includegraphics[width=0.9\textwidth]{Figures/cifar10OSresub.pdf}
 %        \caption{One-shot ResNet-20 CIFAR10}
 %        \label{fig: c10}
 %    \end{subfigure}
 %       \caption{Accuracy versus sparsity for CIFAR 10 and 100}
 %       \label{fig: OS}
%\end{figure}

\begin{table}[ht]%{R}{0.3\textwidth}
  %\begin{center}
    \caption{ResNet-50 on ImageNet sparsity $(\%)$ versus accuracy $(\%)$ results.}\label{tab : res50}
    %\label{tab:table1}
    \setlength{\tabcolsep}{2pt}
    \begin{minipage}{0.5\textwidth}
            \centering
    \begin{tabular}{l c r } % <-- Alignments: 1st column left, 2nd middle and 3rd right, with vertical lines in between
      \textbf{Method} & \textbf{Top-1 Acc} & \textbf{Sparsity}\\
      ResNet-50 & 77.01 & 0\\
      \hline
      %GMP & 75.60 & 80.00 \\
      %DNW & 76.00 & 80.00 \\
      %SNFS $+$ ERK & 75.20 & 80.00 \\
      %RigL $+$ ERK & 75.10 & 80.00 \\
      STR & \textbf{76.19} & 79.55 \\
      STR & 76.12 & \textbf{81.27} \\
      spred\footnotemark & 75.5 & 80.00 \\
      spred & 72.64 & 79.03 \\
      PILoT & 75.62 & 80.00 \\
      \hline
      %GMP & 73.91 & 90.00 \\
      %DNW & 74.00 & 90.00 \\
      %SNFS $+$ ERK & 72.90 & 90.00 \\
      %RigL $+$ ERK & 73.00 & 90.00 \\
      STR & \textbf{74.73} & 87.7 \\
      STR & 74.01 & 90.55 \\
      spred & 71.84 & 89.26 \\
      PILoT & \textbf{74.73} & 88.00 \\
      PILoT & 74.04 & \textbf{91.00}\\
    \end{tabular}
    \end{minipage}
    \begin{minipage}{0.5\textwidth}
            \centering
    \begin{tabular}{l c r } % <-- Alignments: 1st column left, 2nd middle and 3rd right, with vertical lines in between
      \textbf{Method} & \textbf{Top-1 Acc} & \textbf{Sparsity}\\
      ResNet-50 & 77.01 & 0\\
      \hline
      %GMP & 70.59 & 95.00 \\
      %DNW & 68.30 & 95.00 \\
      %RigL $+$ ERK & 70.00 & 95.00 \\
      STR & 70.4 & 95.03 \\
      spred & 69.47 & 94.50 \\
      PILoT & \textbf{72.67} & 94.00\\
      PILoT & \textbf{71.30} & 95.00\\
      PILoT & \textbf{71.05} & \textbf{95.60}\\
      PILoT & \textbf{70.49} & \textbf{96.00}\\
      \hline
      %RigL $+$ ERK & 67.20 & 96.50 \\
      STR & 67.22 & 96.53 \\
      spred & 66.12 & \textbf{97.19}\\
      PILoT & \textbf{68.49} & \textbf{97.19} \\
      \hline
      %GMP & 57.90 & 98.00 \\
      %DNW & 58.20 & 98.00 \\
      STR & 61.46 & 98.05 \\
      spred & 62.71 & 98.20 \\
      PILoT & \textbf{66.49} & 97.75 \\
      PILoT & \textbf{64.06} &\textbf{98.20} \\
    \end{tabular}
    \end{minipage}
  %\end{center}
\end{table}
\footnotetext{Starting from a pretrained model with $77\%$ validation accuracy}
%We have implemented PILoT on ImageNet with ResNet-50 model as in the setting of \citep{kusupati20}. 

In Table \ref{tab : res50}, we compare PILoT to both STR and spred on ImageNet \citep{5206848}. 
See Appendix \ref{appendix : OS section} Table \ref{tab : res50 config} for details on experimental configurations.
Our method competes with or outperforms all baselines at medium and high sparsity levels.
In addition, it improves over spred at $80\%$ sparsity, even when spred is initialized with a $77\%$ pretrained ResNet-50.
This highlights the effectiveness of PILoT and confirms the insights from the developed theory.

\begin{figure}
    \centering
    \includegraphics[width=0.95\textwidth]{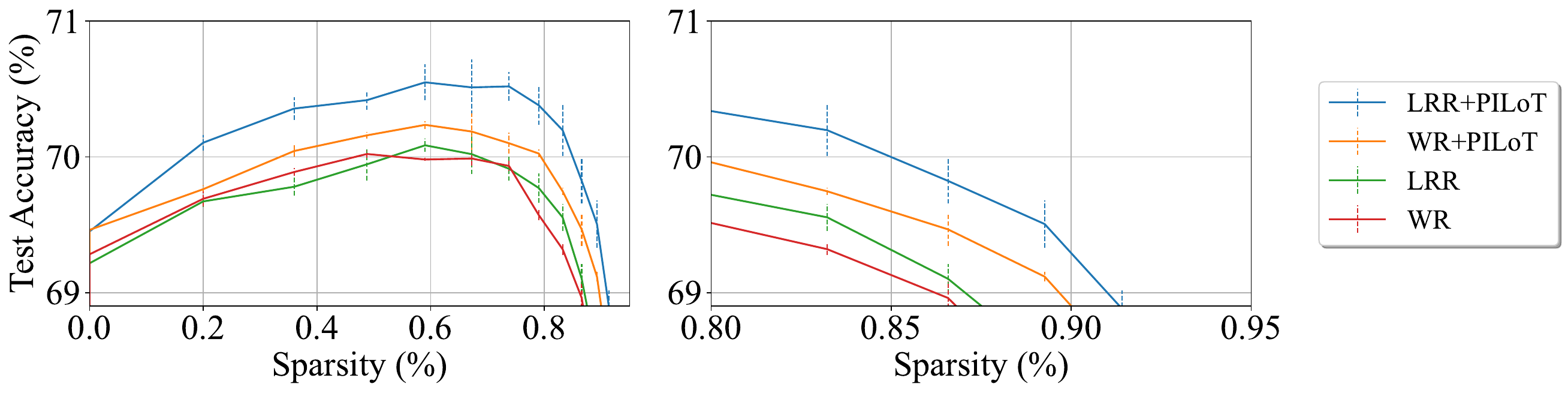}
    \caption{Learning Rate Rewinding (LRR) and Weight Rewinding (WR) with PILoT on ImageNet ResNet-18. The left plot is the complete plot and the right plot is a zoomed-in version.}
    \label{fig: LRR IN}
\end{figure}

\paragraph{Iterative Pruning.}
To demonstrate the versatility of PILoT, we also combine it with the state-of-the-art iterative pruning methods Learning Rate Rewinding (LRR) \citep{maene2021understanding} and Weight Rewinding (WR) \citep{DBLP:journals/corr/abs-1912-05671} on ImageNet using a ResNet-18. 
For simplicity, we use $\beta = 1$ and no regularization.
We see in Figure \ref{fig: LRR IN} that the parameterization $m \odot w$ futher boosts the performance of iterative methods.
Remarkably, WR becomes competitive with LRR.
Additional experiments on CIFAR10 and CIFAR100 with regularization are in Appendix \ref{appendix : itter}.
The regularization further helps to reach higher accuracy at high sparsities.

% to do higher regularization and lower K
% still 80\% has to be better.

%\paragraph{Compression}
%In this experiment we start with a pretrained model and retrain it for sparsity.
%We will train a pretrained ResNet 50 with $77\%$ validation accuracy on ImageNet as in \citep{kusupati20}. 

%\begin{itemize}
%    \item table with str spred and the rest
%\end{itemize}

%Note that spred needs an $80\%$ pretrained model to beat the SOTA, while these are based on a baseline $77\%$.
%In contrast, PILoT is SOTA with the corresponding baseline.

\section{Discussion}
We have shed light on the inner workings of continuous sparsification. 
%Its main challenge is to solve an intractable optimization problem of mixed discrete and continuous nature.
Its basic relaxed formulation utilizes the parameterization $m \odot w$, which induces an implicit bias towards sparsity. 
In contrast to an explicit $L_1$-regularization, it enjoys all the benefits of an implicit regularization that caters first to the loss and not a sparsity penalty.
%This explains why it can often achieve superior performance in the context of deep learning.
%While recent work \citep{ziyin2023spredsolvingl1penalty} has exploited this mechanism, its initialization and static regularization 
%the implicit regularization changes from $L_2$ to $L_1$ during the learning dynamics with the help of explicit dynamic regularization.
Exploiting this insight for neural network sparsification, we have proposed PILoT, which relies on a controllable regularization that acts like an implicit regularization in the original neural network parameter space and, remarkably, corresponds to a time-dependent Bregman potential.
As we have shown, the time-dependent control enables the associated mirror flow to enter the so-called rich regime, and thus effectively change the implicit regularization from $L_2$ to $L_1$. 
This property is central to showing convergence of our approach for (quasi)-convex loss functions and optimality for underdetermined linear regression.
Experiments on standard vision benchmarks further corroborate the utility of our theoretical insights, as our proposal PILoT
achieves significant improvements over state-of-the-art baselines. 

\newpage

\section*{Acknowledgments and Disclosure of Funding}
The authors gratefully acknowledge the Gauss Centre for Supercomputing e.V. for funding this project by providing computing time on the GCS Supercomputer JUWELS at Jülich Supercomputing Centre (JSC). We also gratefully acknowledge funding from the European Research Council (ERC) under the Horizon Europe Framework Programme (HORIZON) for proposal number 101116395 SPARSE-ML.

\bibliography{iclr2025_conference}

\begin{thebibliography}{82}
\providecommand{\natexlab}[1]{#1}
\providecommand{\url}[1]{\texttt{#1}}
\expandafter\ifx\csname urlstyle\endcsname\relax
  \providecommand{\doi}[1]{doi: #1}\else
  \providecommand{\doi}{doi: \begingroup \urlstyle{rm}\Url}\fi

\bibitem[Albarghouthi(2021)]{albarghouthi2021introduction}
Aws Albarghouthi.
\newblock Introduction to neural network verification, 2021.

\bibitem[Alvarez et~al.(2004)Alvarez, Bolte, and Brahic]{Alvarez_2004}
Felipe Alvarez, Jérôme Bolte, and Olivier Brahic.
\newblock Hessian riemannian gradient flows in convex programming.
\newblock \emph{SIAM Journal on Control and Optimization}, 43\penalty0 (2):\penalty0 477–501, January 2004.
\newblock ISSN 1095-7138.
\newblock \doi{10.1137/s0363012902419977}.
\newblock URL \url{http://dx.doi.org/10.1137/S0363012902419977}.

\bibitem[Beck \& Teboulle(2003)Beck and Teboulle]{BECK2003167}
Amir Beck and Marc Teboulle.
\newblock Mirror descent and nonlinear projected subgradient methods for convex optimization.
\newblock \emph{Operations Research Letters}, 31\penalty0 (3):\penalty0 167--175, 2003.
\newblock ISSN 0167-6377.
\newblock \doi{https://doi.org/10.1016/S0167-6377(02)00231-6}.
\newblock URL \url{https://www.sciencedirect.com/science/article/pii/S0167637702002316}.

\bibitem[Boyd \& Vandenberghe(2009)Boyd and Vandenberghe]{Boyd2009ConvexO}
Stephen~P. Boyd and Lieven Vandenberghe.
\newblock Convex optimization.
\newblock 2009.
\newblock URL \url{https://web.stanford.edu/~boyd/cvxbook/}.

\bibitem[Burkholz(2022{\natexlab{a}})]{convexist}
Rebekka Burkholz.
\newblock Convolutional and residual networks provably contain lottery tickets.
\newblock In \emph{International Conference on Machine Learning}, 2022{\natexlab{a}}.

\bibitem[Burkholz(2022{\natexlab{b}})]{depthexist}
Rebekka Burkholz.
\newblock Most activation functions can win the lottery without excessive depth.
\newblock In \emph{Advances in Neural Information Processing Systems}, 2022{\natexlab{b}}.

\bibitem[Burkholz et~al.(2022)Burkholz, Laha, Mukherjee, and Gotovos]{uniExist}
Rebekka Burkholz, Nilanjana Laha, Rajarshi Mukherjee, and Alkis Gotovos.
\newblock On the existence of universal lottery tickets.
\newblock In \emph{International Conference on Learning Representations}, 2022.

\bibitem[Chen et~al.(2022)Chen, Zhang, Wu, Huang, Liu, Chang, and Wang]{chen2022can}
Tianlong Chen, Zhenyu Zhang, Jun Wu, Randy Huang, Sijia Liu, Shiyu Chang, and Zhangyang Wang.
\newblock Can you win everything with a lottery ticket?
\newblock \emph{Transactions on Machine Learning Research}, 2022.
\newblock ISSN 2835-8856.
\newblock URL \url{https://openreview.net/forum?id=JL6MU9XFzW}.

\bibitem[Chizat \& Bach(2020)Chizat and Bach]{pmlr-v125-chizat20a}
L\'ena\"ic Chizat and Francis Bach.
\newblock Implicit bias of gradient descent for wide two-layer neural networks trained with the logistic loss.
\newblock In Jacob Abernethy and Shivani Agarwal (eds.), \emph{Proceedings of Thirty Third Conference on Learning Theory}, volume 125 of \emph{Proceedings of Machine Learning Research}, pp.\  1305--1338. PMLR, 09--12 Jul 2020.
\newblock URL \url{https://proceedings.mlr.press/v125/chizat20a.html}.

\bibitem[Chou et~al.(2024)Chou, Maly, and Stöger]{chou2024induce}
Hung-Hsu Chou, Johannes Maly, and Dominik Stöger.
\newblock How to induce regularization in linear models: A guide to reparametrizing gradient flow, 2024.

\bibitem[da~Cunha et~al.(2022)da~Cunha, Natale, and Viennot]{cnnexist}
Arthur da~Cunha, Emanuele Natale, and Laurent Viennot.
\newblock Proving the lottery ticket hypothesis for convolutional neural networks.
\newblock In \emph{International Conference on Learning Representations}, 2022.

\bibitem[Deng et~al.(2009)Deng, Dong, Socher, Li, Li, and Fei-Fei]{5206848}
Jia Deng, Wei Dong, Richard Socher, Li-Jia Li, Kai Li, and Li~Fei-Fei.
\newblock Imagenet: A large-scale hierarchical image database.
\newblock In \emph{2009 IEEE Conference on Computer Vision and Pattern Recognition}, pp.\  248--255, 2009.
\newblock \doi{10.1109/CVPR.2009.5206848}.

\bibitem[Dereich \& Kassing(2024)Dereich and Kassing]{dereich2024convergence}
Steffen Dereich and Sebastian Kassing.
\newblock Convergence of stochastic gradient descent schemes for lojasiewicz-landscapes, 2024.

\bibitem[Dettmers \& Zettlemoyer(2019)Dettmers and Zettlemoyer]{sparse-momentum-dettmers}
Tim Dettmers and Luke Zettlemoyer.
\newblock Sparse networks from scratch: Faster training without losing performance.
\newblock 2019.

\bibitem[Du et~al.(2017)Du, Jin, Lee, Jordan, Poczos, and Singh]{du2017gradient}
Simon~S. Du, Chi Jin, Jason~D. Lee, Michael~I. Jordan, Barnabas Poczos, and Aarti Singh.
\newblock Gradient descent can take exponential time to escape saddle points, 2017.

\bibitem[Evci et~al.(2020)Evci, Gale, Menick, Castro, and Elsen]{evci-rigl}
Utku Evci, Trevor Gale, Jacob Menick, Pablo~Samuel Castro, and Erich Elsen.
\newblock Rigging the lottery: Making all tickets winners.
\newblock In \emph{International Conference on Machine Learning}, pp.\  2943--2952. PMLR, 2020.

\bibitem[Even et~al.(2023)Even, Pesme, Gunasekar, and Flammarion]{Even2023SGDOD}
Mathieu Even, Scott Pesme, Suriya Gunasekar, and Nicolas Flammarion.
\newblock (s)gd over diagonal linear networks: Implicit regularisation, large stepsizes and edge of stability.
\newblock \emph{ArXiv}, abs/2302.08982, 2023.
\newblock URL \url{https://api.semanticscholar.org/CorpusID:268042036}.

\bibitem[Fawzi et~al.(2022)Fawzi, Balog, Huang, Hubert, Romera-Paredes, Barekatain, Novikov, R.~Ruiz, Schrittwieser, Swirszcz, Silver, Hassabis, and Kohli]{matrix}
Alhussein Fawzi, Matej Balog, Aja Huang, Thomas Hubert, Bernardino Romera-Paredes, Mohammadamin Barekatain, Alexander Novikov, Francisco~J. R.~Ruiz, Julian Schrittwieser, Grzegorz Swirszcz, David Silver, Demis Hassabis, and Pushmeet Kohli.
\newblock Discovering faster matrix multiplication algorithms with reinforcement learning.
\newblock \emph{Nature}, 610\penalty0 (7930):\penalty0 47--53, 2022.

\bibitem[Fehrman et~al.(2019)Fehrman, Gess, and Jentzen]{fehrman2019convergence}
Benjamin Fehrman, Benjamin Gess, and Arnulf Jentzen.
\newblock Convergence rates for the stochastic gradient descent method for non-convex objective functions, 2019.

\bibitem[Ferbach et~al.(2022)Ferbach, Tsirigotis, Gidel, and Avishek]{equivariantLT}
Damien Ferbach, Christos Tsirigotis, Gauthier Gidel, and Bose Avishek.
\newblock A general framework for proving the equivariant strong lottery ticket hypothesis, 2022.

\bibitem[Fischer \& Burkholz(2021)Fischer and Burkholz]{plant}
Jonas Fischer and Rebekka Burkholz.
\newblock Plant 'n' seek: Can you find the winning ticket?, 2021.

\bibitem[Frankle \& Carbin(2018)Frankle and Carbin]{Frankle2018TheLT}
Jonathan Frankle and Michael Carbin.
\newblock The lottery ticket hypothesis: Finding sparse, trainable neural networks.
\newblock \emph{arXiv: Learning}, 2018.
\newblock URL \url{https://api.semanticscholar.org/CorpusID:53388625}.

\bibitem[Frankle \& Carbin(2019)Frankle and Carbin]{frankle2019lottery}
Jonathan Frankle and Michael Carbin.
\newblock The lottery ticket hypothesis: Finding sparse, trainable neural networks.
\newblock In \emph{International Conference on Learning Representations}, 2019.

\bibitem[Frankle et~al.(2019)Frankle, Dziugaite, Roy, and Carbin]{DBLP:journals/corr/abs-1912-05671}
Jonathan Frankle, Gintare~Karolina Dziugaite, Daniel~M. Roy, and Michael Carbin.
\newblock Linear mode connectivity and the lottery ticket hypothesis.
\newblock \emph{CoRR}, abs/1912.05671, 2019.
\newblock URL \url{http://arxiv.org/abs/1912.05671}.

\bibitem[Frankle et~al.(2021)Frankle, Dziugaite, Roy, and Carbin]{frankle2021review}
Jonathan Frankle, Gintare~Karolina Dziugaite, Daniel Roy, and Michael Carbin.
\newblock Pruning neural networks at initialization: Why are we missing the mark?
\newblock In \emph{International Conference on Learning Representations}, 2021.

\bibitem[Gadhikar \& Burkholz(2024)Gadhikar and Burkholz]{gadhikar2024masks}
Advait Gadhikar and Rebekka Burkholz.
\newblock Masks, signs, and learning rate rewinding.
\newblock In \emph{Twelfth International Conference on Learning Representations}, 2024.
\newblock URL \url{https://openreview.net/forum?id=qODvxQ8TXW}.

\bibitem[Gadhikar et~al.(2023)Gadhikar, Mukherjee, and Burkholz]{er-paper}
Advait~Harshal Gadhikar, Sohom Mukherjee, and Rebekka Burkholz.
\newblock Why random pruning is all we need to start sparse.
\newblock In \emph{International Conference on Machine Learning}, 2023.

\bibitem[Gunasekar et~al.(2017)Gunasekar, Woodworth, Bhojanapalli, Neyshabur, and Srebro]{gunasekar2017implicit}
Suriya Gunasekar, Blake Woodworth, Srinadh Bhojanapalli, Behnam Neyshabur, and Nathan Srebro.
\newblock Implicit regularization in matrix factorization, 2017.

\bibitem[Gunasekar et~al.(2020)Gunasekar, Lee, Soudry, and Srebro]{gunasekar2020characterizing}
Suriya Gunasekar, Jason Lee, Daniel Soudry, and Nathan Srebro.
\newblock Characterizing implicit bias in terms of optimization geometry, 2020.

\bibitem[Han et~al.(2015)Han, Pool, Tran, and Dally]{han-efficient-nns}
Song Han, Jeff Pool, John Tran, and William Dally.
\newblock Learning both weights and connections for efficient neural network.
\newblock \emph{Advances in neural information processing systems}, 28, 2015.

\bibitem[Hossain et~al.(2024)Hossain, Fischer, Burkholz, and Quackenbush]{hossain2024tickets}
Intekhab Hossain, Jonas Fischer, Rebekka Burkholz, and John Quackenbush.
\newblock Not all tickets are equal and we know it: Guiding pruning with domain-specific knowledge, 2024.

\bibitem[Jin et~al.(2022)Jin, Carbin, Roy, Frankle, and Dziugaite]{prune-regularize}
Tian Jin, Michael Carbin, Daniel~M. Roy, Jonathan Frankle, and Gintare~Karolina Dziugaite.
\newblock Pruning{\textquoteright}s effect on generalization through the lens of training and regularization.
\newblock In Alice~H. Oh, Alekh Agarwal, Danielle Belgrave, and Kyunghyun Cho (eds.), \emph{Advances in Neural Information Processing Systems}, 2022.
\newblock URL \url{https://openreview.net/forum?id=OrcLKV9sKWp}.

\bibitem[Jumper et~al.(2021)Jumper, Evans, Pritzel, Green, Figurnov, Ronneberger, Tunyasuvunakool, Bates, {\v{Z}}{\'\i}dek, Potapenko, et~al.]{alphafold}
John Jumper, Richard Evans, Alexander Pritzel, Tim Green, Michael Figurnov, Olaf Ronneberger, Kathryn Tunyasuvunakool, Russ Bates, Augustin {\v{Z}}{\'\i}dek, Anna Potapenko, et~al.
\newblock Highly accurate protein structure prediction with {AlphaFold}.
\newblock \emph{Nature}, 596\penalty0 (7873):\penalty0 583--589, 2021.

\bibitem[Kaack et~al.(2022)Kaack, Donti, Strubell, Kamiya, Creutzig, and Rolnick]{kaack2022aligning}
Lynn~H Kaack, Priya~L Donti, Emma Strubell, George Kamiya, Felix Creutzig, and David Rolnick.
\newblock Aligning artificial intelligence with climate change mitigation.
\newblock \emph{Nature Climate Change}, 12\penalty0 (6):\penalty0 518--527, 2022.

\bibitem[Karimi et~al.(2016)Karimi, Nutini, and Schmidt]{10.1007/978-3-319-46128-1_50}
Hamed Karimi, Julie Nutini, and Mark Schmidt.
\newblock Linear convergence of gradient and proximal-gradient methods under the polyak-{\l}ojasiewicz condition.
\newblock In Paolo Frasconi, Niels Landwehr, Giuseppe Manco, and Jilles Vreeken (eds.), \emph{Machine Learning and Knowledge Discovery in Databases}, pp.\  795--811, Cham, 2016. Springer International Publishing.
\newblock ISBN 978-3-319-46128-1.

\bibitem[Kusupati et~al.(2020)Kusupati, Ramanujan, Somani, Wortsman, Jain, Kakade, and Farhadi]{kusupati20}
Aditya Kusupati, Vivek Ramanujan, Raghav Somani, Mitchell Wortsman, Prateek Jain, Sham Kakade, and Ali Farhadi.
\newblock Soft threshold weight reparameterization for learnable sparsity.
\newblock In \emph{Proceedings of the International Conference on Machine Learning}, July 2020.

\bibitem[Kuzmin et~al.(2019)Kuzmin, Nagel, Pitre, Pendyam, Blankevoort, and Welling]{kuzmin2019taxonomy}
Andrey Kuzmin, Markus Nagel, Saurabh Pitre, Sandeep Pendyam, Tijmen Blankevoort, and Max Welling.
\newblock Taxonomy and evaluation of structured compression of convolutional neural networks, 2019.

\bibitem[Kuznedelev et~al.(2023)Kuznedelev, Kurtic, Iofinova, Frantar, Peste, and Alistarh]{kuznedelev2023accurate}
Denis Kuznedelev, Eldar Kurtic, Eugenia Iofinova, Elias Frantar, Alexandra Peste, and Dan Alistarh.
\newblock Accurate neural network pruning requires rethinking sparse optimization, 2023.

\bibitem[Lasby et~al.(2023)Lasby, Golubeva, Evci, Nica, and Ioannou]{struct-dst}
Mike Lasby, Anna Golubeva, Utku Evci, Mihai Nica, and Yani Ioannou.
\newblock Dynamic sparse training with structured sparsity.
\newblock \emph{arXiv preprint arXiv:2305.02299}, 2023.

\bibitem[Lee et~al.(2019)Lee, Ajanthan, and Torr]{snip}
Namhoon Lee, Thalaiyasingam Ajanthan, and Philip H.~S. Torr.
\newblock Snip: single-shot network pruning based on connection sensitivity.
\newblock In \emph{International Conference on Learning Representations}, 2019.

\bibitem[Li et~al.(2021)Li, Nguyen, Hegde, and Wong]{li2021implicitsparseregularizationimpact}
Jiangyuan Li, Thanh~V. Nguyen, Chinmay Hegde, and Raymond K.~W. Wong.
\newblock Implicit sparse regularization: The impact of depth and early stopping, 2021.
\newblock URL \url{https://arxiv.org/abs/2108.05574}.

\bibitem[Li et~al.(2023)Li, Nguyen, Hegde, and Wong]{li2023implicitregularizationgroupsparsity}
Jiangyuan Li, Thanh~V. Nguyen, Chinmay Hegde, and Raymond K.~W. Wong.
\newblock Implicit regularization for group sparsity, 2023.
\newblock URL \url{https://arxiv.org/abs/2301.12540}.

\bibitem[Li et~al.(2022)Li, Wang, Lee, and Arora]{Li2022ImplicitBO}
Zhiyuan Li, Tianhao Wang, Jason~D. Lee, and Sanjeev Arora.
\newblock Implicit bias of gradient descent on reparametrized models: On equivalence to mirror descent.
\newblock \emph{ArXiv}, abs/2207.04036, 2022.
\newblock URL \url{https://api.semanticscholar.org/CorpusID:250407876}.

\bibitem[Liu et~al.(2024)Liu, Zhang, He, Wang, Xiao, Ye, Zhou, Ku, and Hui]{liu2024survey}
Bohan Liu, Zijie Zhang, Peixiong He, Zhensen Wang, Yang Xiao, Ruimeng Ye, Yang Zhou, Wei-Shinn Ku, and Bo~Hui.
\newblock A survey of lottery ticket hypothesis, 2024.

\bibitem[Liu \& Wang(2023)Liu and Wang]{handbook}
Shiwei Liu and Zhangyang Wang.
\newblock Ten lessons we have learned in the new "sparseland": A short handbook for sparse neural network researchers, 2023.

\bibitem[Liu et~al.(2021)Liu, Chen, Chen, Shen, Mocanu, Wang, and Pechenizkiy]{random-pruning-vita}
Shiwei Liu, Tianlong Chen, Xiaohan Chen, Li~Shen, Decebal~Constantin Mocanu, Zhangyang Wang, and Mykola Pechenizkiy.
\newblock The unreasonable effectiveness of random pruning: Return of the most naive baseline for sparse training.
\newblock In \emph{International Conference on Learning Representations}, 2021.

\bibitem[Louizos et~al.(2018)Louizos, Welling, and Kingma]{louizos2018learning}
Christos Louizos, Max Welling, and Diederik~P. Kingma.
\newblock Learning sparse neural networks through $l_0$ regularization, 2018.

\bibitem[Luccioni et~al.(2023)Luccioni, Jernite, and Strubell]{luccioni2023power}
Alexandra~Sasha Luccioni, Yacine Jernite, and Emma Strubell.
\newblock Power hungry processing: Watts driving the cost of ai deployment?
\newblock \emph{arXiv preprint arXiv:2311.16863}, 2023.

\bibitem[Maene et~al.(2021)Maene, Li, and Moens]{maene2021understanding}
Jaron Maene, Mingxiao Li, and Marie-Francine Moens.
\newblock Towards understanding iterative magnitude pruning: Why lottery tickets win, 2021.

\bibitem[Malach et~al.(2020)Malach, Yehudai, Shalev-Schwartz, and Shamir]{malach2020proving}
Eran Malach, Gilad Yehudai, Shai Shalev-Schwartz, and Ohad Shamir.
\newblock Proving the lottery ticket hypothesis: Pruning is all you need.
\newblock In \emph{International Conference on Machine Learning}, 2020.

\bibitem[Narodytska et~al.(2020)Narodytska, Zhang, Gupta, and Walsh]{Narodytska2020In}
Nina Narodytska, Hongce Zhang, Aarti Gupta, and Toby Walsh.
\newblock In search for a sat-friendly binarized neural network architecture.
\newblock In \emph{International Conference on Learning Representations}, 2020.
\newblock URL \url{https://openreview.net/forum?id=SJx-j64FDr}.

\bibitem[Nemirovski \& Yudin(1983)Nemirovski and Yudin]{nemirovskiĭ1983problem}
A.S. Nemirovski and D.B. Yudin.
\newblock \emph{Problem Complexity and Method Efficiency in Optimization}.
\newblock A Wiley-Interscience publication. Wiley, 1983.
\newblock ISBN 9780471103455.
\newblock URL \url{https://books.google.de/books?id=6ULvAAAAMAAJ}.

\bibitem[Orseau et~al.(2020)Orseau, Hutter, and Rivasplata]{orseau2020logarithmic}
Laurent Orseau, Marcus Hutter, and Omar Rivasplata.
\newblock Logarithmic pruning is all you need.
\newblock \emph{Advances in Neural Information Processing Systems}, 33, 2020.

\bibitem[Patil \& Dovrolis(2021)Patil and Dovrolis]{patil2021phew}
Shreyas~Malakarjun Patil and Constantine Dovrolis.
\newblock Phew: Constructing sparse networks that learn fast and generalize well without training data.
\newblock In \emph{International Conference on Machine Learning}, pp.\  8432--8442. PMLR, 2021.

\bibitem[Paul et~al.(2023)Paul, Chen, Larsen, Frankle, Ganguli, and Dziugaite]{paul2023unmasking}
Mansheej Paul, Feng Chen, Brett~W. Larsen, Jonathan Frankle, Surya Ganguli, and Gintare~Karolina Dziugaite.
\newblock Unmasking the lottery ticket hypothesis: What's encoded in a winning ticket's mask?
\newblock In \emph{The Eleventh International Conference on Learning Representations}, 2023.
\newblock URL \url{https://openreview.net/forum?id=xSsW2Am-ukZ}.

\bibitem[Pensia et~al.(2020)Pensia, Rajput, Nagle, Vishwakarma, and Papailiopoulos]{pensia2020optimal}
Ankit Pensia, Shashank Rajput, Alliot Nagle, Harit Vishwakarma, and Dimitris Papailiopoulos.
\newblock Optimal lottery tickets via subset sum: Logarithmic over-parameterization is sufficient.
\newblock In \emph{Advances in Neural Information Processing Systems}, volume~33, pp.\  2599--2610, 2020.

\bibitem[Pesme et~al.(2021)Pesme, Pillaud-Vivien, and Flammarion]{pesme2021implicit}
Scott Pesme, Loucas Pillaud-Vivien, and Nicolas Flammarion.
\newblock Implicit bias of sgd for diagonal linear networks: a provable benefit of stochasticity, 2021.

\bibitem[Peste et~al.(2021)Peste, Iofinova, Vladu, and Alistarh]{peste2021acdc}
Alexandra Peste, Eugenia Iofinova, Adrian Vladu, and Dan Alistarh.
\newblock Ac/dc: Alternating compressed/decompressed training of deep neural networks, 2021.

\bibitem[Pham et~al.(2023)Pham, Liu, Xiang, Le, Wen, Tran-Thanh, et~al.]{pham-paths}
Hoang Pham, Shiwei Liu, Lichuan Xiang, Dung~D Le, Hongkai Wen, Long Tran-Thanh, et~al.
\newblock Towards data-agnostic pruning at initialization: What makes a good sparse mask?
\newblock In \emph{Thirty-seventh Conference on Neural Information Processing Systems}, 2023.

\bibitem[Radhakrishnan et~al.(2021)Radhakrishnan, Belkin, and Uhler]{radhakrishnan2021linear}
Adityanarayanan Radhakrishnan, Mikhail Belkin, and Caroline Uhler.
\newblock Linear convergence of generalized mirror descent with time-dependent mirrors, 2021.

\bibitem[Ramesh et~al.(2022)Ramesh, Dhariwal, Nichol, Chu, and Chen]{dalle2}
Aditya Ramesh, Prafulla Dhariwal, Alex Nichol, Casey Chu, and Mark Chen.
\newblock Hierarchical text-conditional image generation with clip latents.
\newblock \emph{arXiv preprint arXiv:2204.06125}, 2022.

\bibitem[Renda et~al.(2020)Renda, Frankle, and Carbin]{rewindVsFinetune}
Alex Renda, Jonathan Frankle, and Michael Carbin.
\newblock Comparing rewinding and fine-tuning in neural network pruning.
\newblock In \emph{International Conference on Learning Representations}, 2020.

\bibitem[Rockafellar \& Fenchel(1970)Rockafellar and Fenchel]{Rockafellar1970ConvexA}
Tyrrel~R Rockafellar and Werner Fenchel.
\newblock \emph{Convex Analysis}.
\newblock 1970.
\newblock URL \url{https://api.semanticscholar.org/CorpusID:198120397}.

\bibitem[Savarese et~al.(2021)Savarese, Silva, and Maire]{savarese2021winning}
Pedro Savarese, Hugo Silva, and Michael Maire.
\newblock Winning the lottery with continuous sparsification, 2021.

\bibitem[Schwarz et~al.(2021)Schwarz, Jayakumar, Pascanu, Latham, and Teh]{schwarz2021powerpropagation}
Jonathan Schwarz, Siddhant~M. Jayakumar, Razvan Pascanu, Peter~E. Latham, and Yee~Whye Teh.
\newblock Powerpropagation: A sparsity inducing weight reparameterisation, 2021.

\bibitem[Sheen et~al.(2024)Sheen, Chen, Wang, and Zhou]{sheen2024implicit}
Heejune Sheen, Siyu Chen, Tianhao Wang, and Harrison~H. Zhou.
\newblock Implicit regularization of gradient flow on one-layer softmax attention, 2024.

\bibitem[Sreenivasan et~al.(2022)Sreenivasan, Sohn, Yang, Grinde, Nagle, Wang, Xing, Lee, and Papailiopoulos]{rare-gems}
Kartik Sreenivasan, Jy-yong Sohn, Liu Yang, Matthew Grinde, Alliot Nagle, Hongyi Wang, Eric Xing, Kangwook Lee, and Dimitris Papailiopoulos.
\newblock Rare gems: Finding lottery tickets at initialization.
\newblock \emph{Advances in Neural Information Processing Systems}, 35:\penalty0 14529--14540, 2022.

\bibitem[Stephenson et~al.(2019)Stephenson, Shane, Chase, Rowland, Ries, Justice, Zhang, Chan, and Cao]{drugdesign}
Natalie Stephenson, Emily Shane, Jessica Chase, Jason Rowland, David Ries, Nicola Justice, Jie Zhang, Leong Chan, and Renzhi Cao.
\newblock Survey of machine learning techniques in drug discovery.
\newblock \emph{Current drug metabolism}, 20\penalty0 (3):\penalty0 185--193, 2019.

\bibitem[Tanaka et~al.(2020)Tanaka, Kunin, Yamins, and Ganguli]{synflow}
Hidenori Tanaka, Daniel Kunin, Daniel~L. Yamins, and Surya Ganguli.
\newblock Pruning neural networks without any data by iteratively conserving synaptic flow.
\newblock In \emph{Advances in Neural Information Processing Systems}, 2020.

\bibitem[Vaškevičius et~al.(2019)Vaškevičius, Kanade, and Rebeschini]{vaškevičius2019implicitregularizationoptimalsparse}
Tomas Vaškevičius, Varun Kanade, and Patrick Rebeschini.
\newblock Implicit regularization for optimal sparse recovery, 2019.
\newblock URL \url{https://arxiv.org/abs/1909.05122}.

\bibitem[Wang et~al.(2020)Wang, Zhang, and Grosse]{grasp}
Chaoqi Wang, Guodong Zhang, and Roger~B. Grosse.
\newblock Picking winning tickets before training by preserving gradient flow.
\newblock In \emph{International Conference on Learning Representations}, 2020.

\bibitem[Wang et~al.(2023)Wang, Liang, Wang, Wang, Gu, Fang, and Wang]{wang2023searching}
Kun Wang, Yuxuan Liang, Pengkun Wang, Xu~Wang, Pengfei Gu, Junfeng Fang, and Yang Wang.
\newblock Searching lottery tickets in graph neural networks: A dual perspective.
\newblock In \emph{The Eleventh International Conference on Learning Representations}, 2023.
\newblock URL \url{https://openreview.net/forum?id=Dvs-a3aymPe}.

\bibitem[Wen et~al.(2016)Wen, Wu, Wang, Chen, and Li]{NIPS2016_41bfd20a}
Wei Wen, Chunpeng Wu, Yandan Wang, Yiran Chen, and Hai Li.
\newblock Learning structured sparsity in deep neural networks.
\newblock In \emph{Advances in Neural Information Processing Systems}, volume~29, 2016.

\bibitem[Wojtowytsch(2021)]{wojtowytsch2021stochastic}
Stephan Wojtowytsch.
\newblock Stochastic gradient descent with noise of machine learning type. part i: Discrete time analysis, 2021.

\bibitem[Woodworth et~al.(2020)Woodworth, Gunasekar, Lee, Moroshko, Savarese, Golan, Soudry, and Srebro]{woodworth2020kernel}
Blake Woodworth, Suriya Gunasekar, Jason~D. Lee, Edward Moroshko, Pedro Savarese, Itay Golan, Daniel Soudry, and Nathan Srebro.
\newblock Kernel and rich regimes in overparametrized models, 2020.

\bibitem[Wu et~al.(2022)Wu, Raghavendra, Gupta, Acun, Ardalani, Maeng, Chang, Aga, Huang, Bai, et~al.]{wu2022sustainable}
Carole-Jean Wu, Ramya Raghavendra, Udit Gupta, Bilge Acun, Newsha Ardalani, Kiwan Maeng, Gloria Chang, Fiona Aga, Jinshi Huang, Charles Bai, et~al.
\newblock Sustainable ai: Environmental implications, challenges and opportunities.
\newblock \emph{Proceedings of Machine Learning and Systems}, 4:\penalty0 795--813, 2022.

\bibitem[You et~al.(2020)You, Li, Xu, Fu, Wang, Chen, Baraniuk, Wang, and Lin]{earlybird}
Haoran You, Chaojian Li, Pengfei Xu, Yonggan Fu, Yue Wang, Xiaohan Chen, Richard~G. Baraniuk, Zhangyang Wang, and Yingyan Lin.
\newblock Drawing early-bird tickets: Toward more efficient training of deep networks.
\newblock In \emph{International Conference on Learning Representations}, 2020.

\bibitem[Zhao et~al.(2022)Zhao, Yang, and He]{Zhao_2022}
Peng Zhao, Yun Yang, and Qiao-Chu He.
\newblock High-dimensional linear regression via implicit regularization.
\newblock \emph{Biometrika}, 109\penalty0 (4):\penalty0 1033–1046, February 2022.
\newblock ISSN 1464-3510.
\newblock \doi{10.1093/biomet/asac010}.
\newblock URL \url{http://dx.doi.org/10.1093/biomet/asac010}.

\bibitem[Zhou et~al.(2021{\natexlab{a}})Zhou, Zhang, Chen, Diao, and Zhang]{zhou2021efficient}
Xiao Zhou, Weizhong Zhang, Zonghao Chen, Shizhe Diao, and Tong Zhang.
\newblock Efficient neural network training via forward and backward propagation sparsification.
\newblock \emph{Advances in Neural Information Processing Systems}, 34:\penalty0 15216--15229, 2021{\natexlab{a}}.

\bibitem[Zhou et~al.(2021{\natexlab{b}})Zhou, Zhang, Xu, and Zhang]{zhou2021effective}
Xiao Zhou, Weizhong Zhang, Hang Xu, and Tong Zhang.
\newblock Effective sparsification of neural networks with global sparsity constraint.
\newblock In \emph{Proceedings of the IEEE/CVF Conference on Computer Vision and Pattern Recognition}, pp.\  3599--3608, 2021{\natexlab{b}}.

\bibitem[Ziyin(2023)]{Ziyin2023SymmetryIS}
Liu Ziyin.
\newblock Symmetry induces structure and constraint of learning.
\newblock 2023.
\newblock URL \url{https://api.semanticscholar.org/CorpusID:263310669}.

\bibitem[Ziyin \& Wang(2023)Ziyin and Wang]{ziyin2023spredsolvingl1penalty}
Liu Ziyin and Zihao Wang.
\newblock spred: Solving $l_1$ penalty with sgd, 2023.
\newblock URL \url{https://arxiv.org/abs/2210.01212}.

\end{thebibliography}
\bibliographystyle{iclr2025_conference}

\appendix

\section{Mirror flow framework}

In this section we present some known results from the mirror flow framework for completeness.
We derive the Bregman potential associated with $m \odot w$ in Theorem \ref{thm : implicit bias}. Next, we will also show the convergence of the loss and provide optimality guarantees in Theorem \ref{Thm : Arora 4.14}.
In addition, we extend Theorem \ref{Thm : Arora 4.14} with Theorem \ref{Thm : Arora 4.14 improvement}. 
Finally, we give the optimality result for diagonal linear networks in Theorem \ref{exp reg : optimality}

\begin{theorem}\label{thm : implicit bias}
    Let the initialization of $m$ and $w$ satisfy $m_{0,i} > |w_{0,i}|$ for all $i \in [n]$. Then the corresponding mirror function is:
\begin{equation}\label{legendrefunction}
    R(x) :=  \frac{1}{4}\sum_{i = 1}^n x_i \text{arcsinh}\left( \frac{x_i}{2 u_{0,i} v_{0,i}}\right) - \sqrt{x^2_i + 4u_{0,i}^2 v_{0,i}^2} - x_i\text{log}\left(\frac{u_{0,i
}}{v_{0,i}}\right)
\end{equation}
where $u_{0,i} = \frac{m_{0,i} + w_{0,i}}{\sqrt{2}}$ and $v_{0,i} = \frac{m_{0,i} -w_{0,i}}{\sqrt{2}}$.
Furthermore, $R$ is a Bregman function.
\end{theorem}
Proof.
The result follows directly from applying Theorem 4.16 in \citep{Li2022ImplicitBO}. 

Theorem \ref{thm : implicit bias} implies the following:
a) The global minima of $R$ is at the initialization $x_0 = m_0 \odot w_0$.
b) The Lipschitz coeficient of $R$ depends on the initalization. The Lipschitz coeficient $L_R$ of (\ref{legendrefunction}) is $L_R = \frac{1}{\min_i {2u_{0,i}v_{0,i}}}$, determining the smoothness of the potential.  
% \begin{itemize}
%     \item The global minima of $R$ is at the initialization $x_0 = m_0 \odot w_0$.
%     \item The Lipschitz coeficient of $R$ depends on the initalization. The Lipschitz coeficient $L_R$ of \ref{legendrefunction} is $L_R = \frac{1}{\min_i {2u_{0,i}v_{0,i}}}$, determining the smoothness of the potential.  
% \end{itemize}
%
Following these two observations we make the following remark about Theorem \ref{thm : implicit bias}.

\begin{remark}\label{remark : Hypentr}
    Note that when the initialization is zero, i.e., $w_{0} = 0, m_0 = \sqrt{a}$ with $a \geq 0$ then (\ref{legendrefunction}) is the hyperbolic entropy. The hyperbolic entropy is
    \begin{equation*}
        \sum_{i = 1}^n x_i \text{arcsinh}\left( \frac{x_i}{a}\right) - \sqrt{x^2_i + a^2}
    \end{equation*}
    Theorem 2 of \citep{woodworth2020kernel} characterizes the behavior in the limit for this case. 
    For the hyperbolic entropy in case $a\rightarrow 0$ and $|\frac{x}{a}| \rightarrow \infty$,
    \begin{equation*}
        R(x) \sim \text{log}\left(\frac{1}{a}\right)||x||_{L_1}.
    \end{equation*}
    This means an $L_1$ bias is induced when $a$ is small. Nevertheless, we need an exponentially small $a$ compared to $x$ to get there as shown in \citep{woodworth2020kernel}, which can lead to numerical problems.
    Furthermore, $m_0 = w_0 = 0$ is a saddle point which can slow down training (exponentially) \citep{du2017gradient}.
    Additionally, the asymptotic result only holds for initializing at zero. Note $L_R = a$ in this case.
\end{remark}
Remark \ref{remark : Hypentr} shows the potential of using the implicit bias to induce sparsity. 
To actualize this, we need to solve the two challenges posed in the remark. 
Both are remedied in Section \ref{section: Theory}.

%\begin{remark}
%    The spred initalization is $w_{0,i} = \text{sign}(x_{0,i}) \sqrt{|x_{0,i}}$ and $m_{0,i} = \sqrt{|x_{0,i}}$ for all $i \in [n]$. Therefore, not covered by Theorem \ref{thm : implicit bias}. The resulting Bregman potential here is given by
%    \begin{equation*}
%        \tilde{R}(x) : = \sum_{i = 1}^n x_i log \left( \frac{x_i}{x_{0,i}}\right) -x_i
%    \end{equation*}
% This implies that their initalization restricts the gradient flow dynamics to either of the half spaces $\mathbb{R}_{\geq}$ or $\mathbb{R}_{\leq}$ depending on the initalization. 
% Therefore, implying the gradient flow cannot flip signs.
% This has been shown to lead to hampering performance.
%\end{remark}

In addition to this promising formulation of implicitly minimizing an $L_1$ norm with the use of the mirror framework, we can get convergence results. These results make it clear why implicit regularization is preferable over explicit regularization. 
The convergence result from \citep{Li2022ImplicitBO} is stated for our setting. Furthermore, the theorem is extended for a specific class of Bregman functions.
\begin{theorem}(Theorem 4.14 \citep{Li2022ImplicitBO})\label{Thm : Arora 4.14}
    Assume that $f$ is quasi-convex, $\nabla f$ is locally Lipschitz and $\text{argmin} \{ f(x) | x \in \mathbb{R}^n\}$ is non-empty. Then as $t \rightarrow \infty$, $x_t$ converges to some critical point $x^*$. Moreover, if $f$ is convex $x_t$ converges to a minimizer of $f$.
\end{theorem}

In Theorem \ref{Thm : Arora 4.14} it is shown that with implicit regularization an optimal solution to the original optimization problem can be reached. 
In contrast, explicit regularization makes this not possible, by definition. 
Because the optimization problem has fundamentally changed. Showing the benefit of implicit over explicit.

For the extension, the convexity constraint is replaced by the Polyak-\L ojasiewicz (PL) inequality in the theorem. 
The PL-inequality is a more realistic constraint in a machine learning context as loss functions are not locally convex but can satisfy the PL inequality locally \citep{wojtowytsch2021stochastic, dereich2024convergence}. 
The PL-inequality for a continuously differentiable function $f$ is
\begin{equation}\label{PL inequality}
    ||\nabla f(x) ||^2_{L_2} \geq \lambda \left(f(x) -f(x^*)\right) \qquad \forall x \in \mathbb{R}^n
\end{equation}
for some $\lambda > 0$ and global minima $x^*$ of $f$. 
This allows us to state the modified theorem.
\begin{theorem}\label{Thm : Arora 4.14 improvement}
    Consider the same setting as Theorem \ref{Thm : Arora 4.14}. Assume $R$ satisfies for all $x \in \mathbb{R}^n$,
    \begin{equation}\label{Inv Hessian Legendre}
    z^T \left(\nabla^2 R(x)\right)^{-1}  z \geq \sigma ||z||^2_{L_2} \qquad \forall z \in \mathbb{R}^n.
\end{equation}
Furthermore, assume $f$ satisfies the PL-inequality (\ref{PL inequality}). Then $x_t$ converges to a minimizer of $f$. Furthermore, the loss converges linearly with rate $\sigma \lambda$.
\end{theorem}
Proof.
The evolution of $f(x_t) -f(x^*)$ is described by $d f(x_t) = - \nabla f(x_t)^{\top} \left(\nabla^2 R(x_t)\right)^{-1} \nabla f(x_t) dt$. 
% \begin{equation*}
%     d f(x_t) = - \nabla f(x_t)^{\top} \left(\nabla^2 R(x_t)\right)^{-1} \nabla f(x_t) dt.
% \end{equation*}
From (\ref{Inv Hessian Legendre}) and (\ref{PL inequality}) the evolution is bounded by
\begin{align*}
    d f(x_t) &\leq - \sigma ||\nabla f(x_t)||^2_{L_2} dt
    \leq -\sigma \lambda  \left(f(x_t) -f (x^*)\right) dt.
\end{align*}
Applying Gronwall's Lemma concludes the proof. $\square$

Note that Theorem \ref{Thm : Arora 4.14 improvement} holds in the same (general) setting as Theorem \ref{Thm : Arora 4.14}. Also, note that the PL-inequality together with quasi-convexity does not imply convexity.
Theorem \ref{Thm : Arora 4.14 improvement} holds for our setting. In this case, it follows from a direct computation that
\begin{equation*}
    \left(\nabla^2 R(x)\right)^{-1} = \text{diag}\left(\sqrt{x^2 + 4u_{0,1}^2 v_{0,1}^2 }, \hdots ,\sqrt{x^2 + 4u_{0,n}^2 v_{0,n}^2 }\right).
\end{equation*}
This implies that $\sigma = 2\min_i {u_{0,i}v_{0,i}}$ in Theorem \ref{Thm : Arora 4.14 improvement}, which again highlights the importance of the initialization. 

Finally, in the case of under-determined linear regression, we can derive optimality conditions in the form of KKT conditions of $R$. Consider a data set $\left(z_j, y_j\right)_{i=1}^d$ with $z_j \in \mathbb{R}^n$ and $y_j \in \mathbb{R}$. Let $Z = \left(z_1 , \hdots z_d\right)$ and $Y = \left( y_1, \hdots y_d \right) $. For the regression to be called underdetermined $n > d$. 

\begin{theorem}\label{exp reg : optimality}(Theorem 4.17 \citep{Li2022ImplicitBO})
    In case of under-determined regression consider the loss function $f(x) = \Tilde{f}(Zx - Y)$. Assume $f$ satisfies the conditions of Theorem \ref{Thm : Arora 4.14}. Then $x_t$ converges to $x^*$ such that
    \begin{equation*}\label{exp reg : opt L1}
        x^* = \text{argmin}_{Zx = Y} R(x)
    \end{equation*}
\end{theorem}
Note that Theorem 4.17 of \citep{Li2022ImplicitBO} only uses quasi-convexity of the loss.
Theorem \ref{exp reg : optimality}  guarantees that the optimization problem is solved while implicitly minimizing the potential $R$.
Thus choosing the sparsest model out of the models that predict the data perfectly.
This highlights another benefit of implicit regularization over explicit regularization.

In this section, we have shown the viability of using the implicit bias framework to induce an implicit regularization. 
Furthermore, we have given two known benefits of using the implicit bias framework over explicit regularization. 
The benefits are convergence to the optimal solution of the original problem and optimality in the case of underdetermined regression.
To add to this, we have extended the convergence theorem using the PL-inequality in \ref{PL inequality}. 
Moreover, we again highlight the importance of the initialization of $m_0$ and $w_0$ with the influence on the smoothness of the Bregman potential and convergence of the loss.
The initialization insight as in the main text is used to improve upon spred \citep{Ziyin2023SymmetryIS} as their initialization has scaling $2u_0v_0 =0$, it follows already from the mirror flow framework that $2u_0v_0 = 1$ is a better initialization.
Furthermore, we also do not initialize at zero though, and scaling needs to be exponentially small to get a good approximation of the $L_1$ norm potentially making it hard to escape the saddle point. 
Therefore, the explicit regularization analyzed in Section \ref{section: Theory} is necessary to exploit the implicit bias framework.

\section{Proof main result}
We show the main result here. 
The proof consists of four parts
\begin{itemize}
    \item $R_{a_t}$ satisfies a mirror flow (Lemmas \ref{appendix : lemma flow} and \ref{appendix : lemma mirror})
    \item Boundedness of the iterates and convergence to a critical point (Lemma \ref{appendix : itter bound})
    \item Convergence of the loss (Theorem \ref{thm : convergence})
    \item Optimality in case of underdetermined linear regression (Theorem \ref{appendix : optimality})
\end{itemize}

Consider the following gradient flow

\begin{equation}\label{appendix : flow mw reg}
\begin{cases}
	d m_t = -\nabla f\left(m_t \odot w_t\right) \odot w_t - 2 \alpha_t m_t dt \\
	d w_t = -\nabla f\left(m_t \odot w_t\right) \odot m_t - 2 \alpha_t w_t dt
\end{cases}
\end{equation}
For the flow in (\ref{appendix : flow mw reg}) to be well-posed $\nabla f$ needs to be locally Lipschitz continuous. This is a sufficient condition given that $\alpha_t$ is "nice", which will be made more rigorous later. The evolution of $x_t = m_t \odot w_t$ is derived in Lemma \ref{appendix : lemma flow}.
\begin{lemma}\label{appendix : lemma flow}
	The evolution of $x_t = m_t \odot w_t$ with \ref{appendix : flow mw reg} is described by
	\begin{equation*}
		x_t = u_0^2 \odot\text{exp}\left( -2\int_0^t \nabla f\left(x_s\right) ds -4 \int_0^t \alpha_s ds\right) - v_0^2\odot\text{exp}\left( 2\int_0^t \nabla f\left(x_s\right) ds - 4\int_0^t \alpha_s ds\right),
	\end{equation*}
where $u_0 = \frac{m_0 + w_0}{\sqrt{2}}$ and $v_0 = \frac{m_0 -w_0}{\sqrt{2}}$.
\end{lemma}
Proof.
This follows from deriving the flow of $m_t$ and $w_t$ and then combining the two. The evolution of both are given by
\begin{equation*}
\begin{cases}
	m_t = \left(m_0 \odot \text{cosh}\left( -\int_0^t \nabla f\left(x_s\right) ds\right) + w_0 \odot \text{sinh}\left( -\int_0^t \nabla f\left(x_s\right) ds \right) \right)  \text{exp}\left(-2 \int_0^t \alpha_s ds\right) \\
	w_t = \left(w_0 \odot \text{cosh}\left( -\int_0^t \nabla f\left(x_s\right) ds\right) + m_0 \odot \text{sinh}\left( -\int_0^t \nabla f\left(x_s\right) ds \right) \right)  \text{exp}\left(-2 \int_0^t \alpha_s ds\right).
\end{cases}
\end{equation*}
For ease of notation set $L_t = \int_0^t \nabla f\left(x_s
\right) ds$ and $A_t = \int_0^t \alpha_s ds$. Combining gives us
\begin{align*}
	x_t &=m_t \odot  w_t \\
 &=  \left(m_0^2 + w_0^2\right) \odot \text{cosh}\left( -L_t\right) \odot  \text{sinh}\left( -L_t\right) \text{exp}\left(-4 A_t\right) \\ &+  w_0 \odot m_0 \odot \left( \text{cosh}\left( -L_t\right)^2 + \text{cosh}\left( -L_t\right)^2\right) \text{exp}\left(-4 A_t\right) \\
&= \left( \frac{\left(m_0^2 + w_0^2\right)}{2} \odot \text{sinh}\left( -2L_t\right)\right)\text{exp}\left(-4A_t\right) \\ &+ w_0 \odot m_0 \odot \left(\text{cosh}\left( -2L_t\right) \right) \text{exp}\left(-4A_t\right) \\
&= u_0^2 \odot \text{exp}\left( -2 L_t -4A_t\right) - v_0^2 \odot \text{exp}\left( 2L_t - 4A_t\right)
\end{align*}
where the second equality follows from hyperbolic identities. $\square$

It follows from Lemma \ref{appendix : lemma flow} and the local Lipschitz condition on $\nabla f$ and $\int_0^t \alpha_s ds < \infty$ for all $t \geq 0$, that the flow is well-posed.
We now define the (corrected)-hyperbolic entropy function. The corrected hyperbolic entropy is given by
\begin{equation*}\label{appendix : hypen}
	R_a (x) = \frac{1}{2}\sum_{i = 1}^n x_i \text{arcsinh}\left( \frac{x_i}{ a} \right) - \sqrt{x_i^2 + a^2} - x_i \text{log} \frac{u_{0 i}}{v_{0, i}},
\end{equation*}
where the last term is the correction. The correction stems from not initializing at zero. 

\begin{lemma}\label{appendix : lemma mirror}
Let $|w_{i 0}| \leq m_{0 i}$ for all $i \in [n]$, then $R_{a_t}(x_t)$ with $a_t = 2u_0 \odot v_0 \text{exp}\left(- 2\int_0^t \alpha_s ds\right)$ satisfies
\begin{equation}\label{appendix : Mirror flow}
	d \nabla R_{a_t} (x_t) = -\nabla f(x_t) dt \qquad x_0 = m_0 \odot w_0.
\end{equation}
\end{lemma}
Proof.
This follows from Lemma \ref{appendix : lemma flow},  
\begin{align*}
	x_t \text{exp}\left(4\int_0^t \alpha_s ds\right) =  u_0^2\text{exp}\left( -2\int_0^t \nabla f(x_s) ds\right) - v_0^2\text{exp}\left( 2\int_0^t \nabla f(x_s) ds\right) \Leftrightarrow \\
\frac{1}{2}\left(\text{arcsinh}(\frac{x_t}{a_t}) - \text{log}(\frac{u_0}{v_0})\right) = - \int_0^t \nabla f(x_s) ds.
\end{align*}
This equivalence follows from setting $z = \text{exp}\left( -2\int_0^t \nabla f(x_s) ds\right)$ and solving the resulting quadratic equation. Notice that the left hand side in (\ref{appendix : Mirror flow}) is $\nabla R_{a_t}(x_t)$. $\square$

\begin{lemma}\label{appendix : itter bound}
    Let $f$ be a quasi-convex function and $\alpha_t \geq 0$ for all $t\geq 0$. Furthermore, assume the integral $\int_0^t \alpha_s ds < \infty$. Then the iterates are bounded and converge to a critical point.
\end{lemma}

Consider the time-dependent Bregman divergence 
\begin{equation*}
     D_{a_t}(x^*, x_t) := R_{a_t}(x^*) - R_{a_t}(x_t) - \nabla_x R_{a_t}^T (x^* - x_t) \geq 0
\end{equation*}
The divergence is bounded by:
\begin{align*}
    D_{a_t}(x^*, x_t) \leq R_{a_{\infty}}(x^*) - R_{a_t}(x_t) - \nabla_x R_{a_t}^T (x^* - x_t) =: W_t,
\end{align*}
this follows from the fact that the map $a \rightarrow R_a$ is decreasing. 
We make the following two observations:
\begin{equation}\label{appendix : observation}
	\frac{d}{d a} R_{a}(x) = -\frac{1}{2}\dfrac{x^2\left|a\right|+a^3}{a^2\sqrt{a^2+x^2}} \leq 0 \qquad \text{ and } \qquad\frac{d a_t}{dt} \leq 0 \qquad \forall t\geq 0.
\end{equation}

This allows us to bound the evolution
\begin{align*}
    \frac{d}{dt} W_t & = \frac{d}{dt}\left(-R_{a_t}\left(x_t\right) - \nabla_x R_{a_t}^T \left(x^* - x_t\right)\right) \\
    &= -\frac{d}{d a}R_{a_t}\left(x_t\right) \frac{d}{dt} a_t  - \frac{d}{dt}\left(  \nabla_x R_{a_t}^T\right) \left(x^* -x_t\right) \\
    &\leq \nabla_x f(x_t)^T  \left(x^* -x_t\right)  \\
    &\leq 0,
\end{align*}
where the observations in (\ref{appendix : observation}) are used in the first inequality.

From Theorem 4.16 in \citep{Li2022ImplicitBO} it follows that for all $a > 0$, $R_a$ is a Bregman potential. 
Implying that for all $a > 0$ the level set for $\gamma \in \mathbb{R}$, 
\begin{equation*}
    \{x \in \mathbb{R}^n : D_{a}(x^*, x) \leq \gamma \}
\end{equation*}
is bounded. 
Combining this with the fact that the evolution is bounded, implies the iterates are bounded. 

In the next part, it is shown that $x_t$ converges to a critical point.
We first show that the loss becomes eventually non-increasing.
There is a $T$ such that for all $t \geq T$ the loss $f$ is non-increasing.,
\begin{align*}
    d f(x_t) &=- \left(\nabla f(x_t)^T \text{diag}\left(\sqrt{x_t^2 + a_t^2}\right) \nabla f(x_t) + 2 \alpha_t \nabla f(x_t)^T x_t \right) dt \\
    &\leq \left(-\nabla f(x_t)^T \text{diag}\left(\sqrt{x_t^2 + a_t^2}\right) \nabla f(x_t) + 2 \alpha_t C \right) dt,
\end{align*}
where it is used that the iterates are bounded and $\nabla f$ is locally Lipschitz. As $t \rightarrow \infty$ we have that $\alpha_t \rightarrow 0$ and $a_t \rightarrow a_{\infty} >0$ by assumption. Hence there exists a $T$ such that for all $t \geq T$ we have 
\begin{equation*}
    d f(x_t) \leq 0.
\end{equation*}
Note if this is not the case then there is a $T > 0$ such that $\nabla f(x_T) = 0$ implying convergence (to a critical point).

Now let $x_{\infty}$ be an accumulation point of the bounded flow $x_t$.
We use this to show convergence to a critical point.
We have that for all $x \in \mathbb{R}^n$,
\begin{equation*}
    \nabla f(x_{\infty})^{\top} x = \lim_{t \rightarrow \infty}  \frac{1}{t}\left(\int_T^{T+t}\nabla f(x_s)ds\right)^{\top} x = \lim_{t \rightarrow \infty}\frac{1}{t}\left(R_{a_T}(x_T) - R_{a_{T + t}}(x_{T+t})\right)^{\top}x = 0,
\end{equation*}
where the first equality follows from that the loss is non-increasing and the second one from the time-dependent mirror flow description. Finally, because $x_t$ converges to an accumulation point we also have $\lim_{t \rightarrow \infty} R_{a_{t}}(x_{t}) = R_{a_{\infty}}(x_{\infty})$ by continuity, giving the last equality.

We use that the accumulation point is a critical point and set $x^* = x_{\infty}$ in $W_t$ such that $W_t \rightarrow 0$.
This implies $D_{a_t}(x_{\infty}, x_t) \rightarrow 0$ by the upperbound.
It follows from the fact that the iterates are bounded that $R_{a_t}$ is $\mu$-strongly convex on this bounded convex set where the iterates stay. 
This gives
\begin{equation*}
    ||x_{\infty} - x_t ||_{L_2} \leq \frac{\mu}{2} D_{a_t}(x_{\infty}, x_t) \rightarrow 0,
\end{equation*}
showing $x_t$ converges to a critical point.
$\square$

Lemma \ref{appendix : itter bound} gives a condition such that the iterates are bounded and converge to a critical point. 
It remains to be shown that the loss converges.
This is done in Theorem \ref{thm : convergence}.
\begin{theorem}\label{thm : convergence}
    Consider the same setting as Lemma \ref{appendix : itter bound}, if $f$ is convex or satisfies the PL-inequality we have convergence to an interpolator $x^*$ such that it is a minimizer of $f$. Furthermore, in the PL-inequality case, the loss converges linearly. 
\end{theorem}
Proof.
Assume $f$ is convex, notice first that there is a $T$ such that for all $t \geq T$ the loss is non-increasing. 
Combining this with a bound on the time-dependent Bregman potential gives us convergence of the loss. 
The time-dependent Bregman divergence is again defined by
\begin{equation*}
    D_{a_t}(x^*, x_t) = R_{a_t}(x^*) - R_{a_t}(x_t) - \nabla_x R_{a_t}^{\top} (x^* - x_t) \geq 0.
\end{equation*}
The divergence is bounded by:
\begin{align*}
    D_{a_t}(x^*, x_t) \leq W_t.
\end{align*}
The evolution of the bound is
\begin{align*}
    \frac{d}{dt} W_t & = \frac{d}{dt}\left(-R_{a_t}\left(x_t\right) - \nabla_x R_{a_t}^{\top} \left(x^* - x_t\right)\right) \\
    &= -\frac{d}{d a}R_{a_t}\left(x_t\right) \frac{d}{dt} a_t  - \frac{d}{dt}\left(  \nabla_x R_{a_t}^{\top}\right) \left(x^* -x_t\right) \\
    &\leq \nabla_x f(x_t)^{\top}  \left(x^* -x_t\right)  \\
    &\leq f\left(x^*\right) - f\left(x_t\right),
\end{align*}
where again the observations in (\ref{appendix : observation}) are used in the first inequality.
Therefore the loss converges:
\begin{align*}
    f(x_{T+t}) - f(x^*) &\leq \frac{1}{t} \int_T^{T+t} f(x_s) - f(x^*) ds \\
    &\leq \frac{W_T - W_{T+t} }{t} \\
    &\leq \frac{W_T}{t} \rightarrow 0
\end{align*}
where the first inequality follows from convexity of the loss and the third inequality from the fact that $W_t \geq D_{a_t}(x^*,x_t) \geq 0$.  So the loss converges. We already know from Lemma \ref{appendix : itter bound} that the iterates converge, concluding the convex case.

In case when $f$ satisfies the PL-inequality, we proceed in the same way as Theorem \ref{Thm : Arora 4.14 improvement}. The evolution of $f$ is given by
\begin{align*}
    d f(x_t) &=- \left(\nabla f(x_t)^{\top} \text{diag}\left(\sqrt{x_t^2 + a_t^2}\right) \nabla f(x_t) + 2 \alpha_t \nabla f(x_t)^{\top} x_t \right) dt \\
    &\leq \left(- A_{\infty} \lambda\left( f(x_t) -f(x^*)\right)+ \alpha_t C ||x^*||_{L_2} \right)dt
\end{align*}
where $C$ is constant depending on the smoothness of $f$ and $A_{\infty} = 2\min_i u_{0,i} v_{0,i} \exp \left( -2\int_0^{\infty} \alpha_s ds\right)$. Then it follows from Gronwall's Lemma that
\begin{equation*}
    f(x_t) -f(x^*) \leq \left(f(x_0) -f(x^*)\right)\text{exp}\left(-A_{\infty} \lambda t + \int_0^t \alpha_s C||x^*||_{L_2} ds\right).
\end{equation*}
It follows from the fact that $\int_0^t \alpha_s ds < \infty$ for all $t \geq 0$ that the loss $f$ convergence. 
Convergence of the iterates now follows in a similar way as the convex case. $\square$

We now show optimality in the case of underdetermined linear regression
Consider a data set $\left(z_j, y_j\right)_{i=1}^d$ with $z_j \in \mathbb{R}^n$ and $y_j \in \mathbb{R}$. Let $Z = \left(z_1 , \hdots z_d\right)$ and $Y = \left( y_1, \hdots y_d \right) $. For the regression to be called underdetermined $n > d$.

\begin{theorem}\label{appendix : optimality}
    In case of under-determined regression consider the loss function $f(x) = \Tilde{f}(Zx - Y)$. Assume $f$ satisfies the conditions with at least one of the convergence criteria of Theorem \ref{thm : convergence}. Then $x_t$ converges to $x^*$ such that
    \begin{equation}\label{appendix : opt L1}
        x^* = \text{argmin}_{Zx = Y} R_{a_{\infty}}(x)
    \end{equation}
\end{theorem}
Proof.
Convergence follows from Theorem \ref{thm : convergence}. It remains to be shown that the optimality conditions of (\ref{appendix : opt L1}) are satisfied. The gradient flow of $R_{a_t}$ satisfies
\begin{equation*}
    \nabla R_{a_{t}}(x_{t}) = Z^{\top} \int_0^{t} \nabla \Tilde{f}(x_s) ds \in \text{span}\{Z^{\top}\}.
\end{equation*}
This quantity is well defined for all $t \geq 0$ because $\nabla \tilde{f}$ is locally Lipschitz (because $f$ has to be locally Lipschitz). Therefore taking the $t \rightarrow \infty$ yields the KKT conditions of the optimization problem in (\ref{appendix : opt L1}). $\square$

\subsection{Discussion of the proof}
Most of the proof follows the same arguments as in \citep{Alvarez_2004, Li2022ImplicitBO, pesme2021implicit}. 
The notable differences are showing that the loss becomes decreasing over time and the observations made in (\ref{appendix : observation}). 

\section{Details experiments}\label{appendix : details experiments}
In this section, we provide the details of the experiments. 
In addition, there are additional figures given.

\paragraph*{Compute}
The codebase for the experiments is written in PyTorch and torchvision and their relevant primitives for model construction and data-related operations. The experiments in the paper are trained on an NVIDIA A6000. In addition, the diagonal linear network is trained on a CPU 13th Gen INTEL(R) Core(TM) i9-13900H.

\subsection{Diagonal linear network}\label{dln appenix}
For each setting, different regularization schemes are tried. 
In total, $7$ options are tried.  
$4$ of the schedules are constant i.e. the regularization stays the same during training. The remaining $3$ are decaying schedules. These schedules we name harmonic, quadratic, and geometric. The schedules are described by the following recurrent relations:
\begin{equation*}
    h_k = \frac{1}{k} \qquad \text{ , } \qquad q_k = \frac{1}{k^2} \qquad \text{ and } \qquad g_k = p^k
\end{equation*}
where we have set $p = 0.95$. These schedules lead to a total strength of regularization applied. We denote $A := \int_0^t \alpha_s ds$ the total strength of the regularization. So in practice, it is the weighted sum of the regularization strength. 

\begin{figure}[ht]
    \centering
    \includegraphics[width=\textwidth]{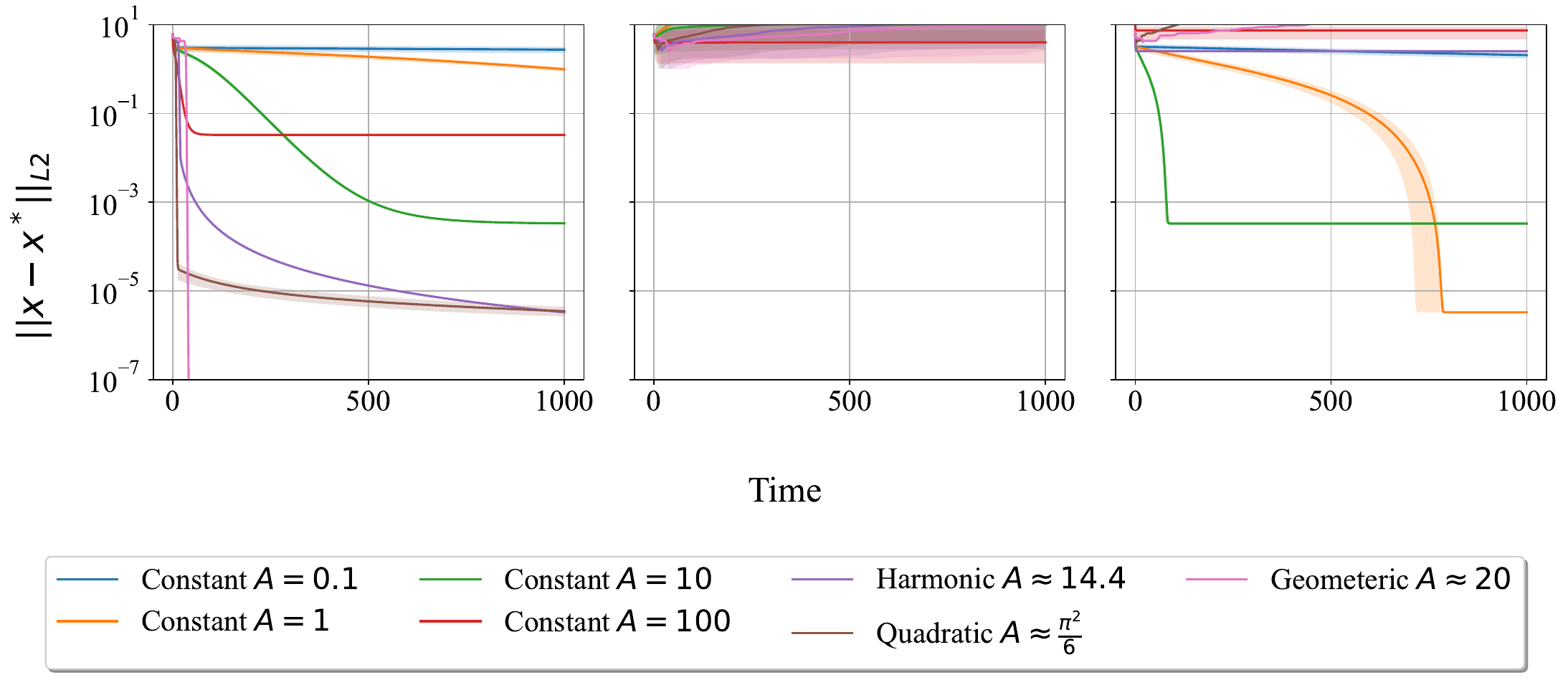}
         \caption{All runs for the diagonal linear network. From left to right $m \odot w$ with PILoT initalization, $m \odot w$ with spred initialization, and $x$ with $L_1$ regularization}
         \label{fig: appendix all runs}
\end{figure}

In Figure \ref{fig: appendix all runs} we present the trajectories of all the schedules for the $3$ considered settings. We observe that our regularization performs the best with the decaying schedules as predicted by the theory. The other methods need constant regularization to perform well as already mentioned in Remark \ref{exp reg : remark reg}. Note that, PILoT also can perform well with constant regularization (see Figure \ref{fig: appendix all runs}.

\subsection{One-shot}\label{appendix : OS section}
In this section we give the additional details for the one-shot experiments.
In Table \ref{experiment : oneshot} the hyperparameters for the CIFAR 10 and 100 experiments are given. To determine which configuration is best for which sparsity level we compute the validation accuracy at multiple levels and choose the level just before the accuracy drops $1\%$ or in the high-sparsity regime $2\%$. 
Moreover, we use $s = -200$ for STR.

For the ImageNet experiment we use the setup of \citep{kusupati20}. 
For PILoT and spred we in addition use $L_2$ regularization to compensate for the weight decay i.e. we add a term $\left(m\odot w\right)^2$ with strength $0.000030517578125/2$, which is based on the weight decay strength in \citep{kusupati20} for the baseline. 
Furthermore, weight decay is turned off for the other parameters.
In Table \ref{tab : res50 config} we present the configurations that correspond to the values in Table \ref{tab : res50}.
Note for all PILoT configs $\delta = 1.01$ is used.

\begin{table}[ht]
\setlength{\abovecaptionskip}{5pt}
  \caption{One-shot experiment} 
  \label{experiment : oneshot}
  \centering
  \begin{tabular}{lll}
    \toprule
    %\multicolumn{2}{c}{Part}                   \\
    \cmidrule(r){1-2}
    Parameter     & Setting   & Comments  \\
    \midrule
    Optimizer &  SGD   &  \\
    Momentum & 0.9 & \\
    Batch size & 256 & \\
    Activation function & ReLu & \\
    Weight decay\footnotemark & $10^{-4}$ & \\
    Base learning rate & $\{0.1, 0.2 \}$ &  \\ 
    Epochs & $150$ & \\
    Warmup period & $0$ & \\
    Initialization & Kaiming normal & \\
    Scaling & $1$ & Only for $m \odot w$ \\
    $\delta$ & $1.01$ & \\
    $K$ & $8000$ & \\
    \midrule
    CIFAR 10 & & \\
    \midrule
    Learning rate schedule    & cosine warmup  \\
    \midrule
    CIFAR 100 & & \\
    \midrule
    Learning rate schedule    & step warmup  \\
    \bottomrule
    \end{tabular}
\end{table}
\footnotetext{Applied to the other parameters}

\begin{table}[ht]%{R}{0.3\textwidth}
  %\begin{center}
    \caption{ResNet-50 on ImageNet configurations for each sparsity $(\%)$.}\label{tab : res50 config}
    %\label{tab:table1}
    \setlength{\tabcolsep}{2pt}
    \begin{minipage}{0.5\textwidth}
            \centering
    \begin{tabular}{l c c r } % <-- Alignments: 1st column left, 2nd middle and 3rd right, with vertical lines in between
      \textbf{Method} & $\alpha_{init}$ & $K$ & sparsity\\
      \hline
      %GMP & 75.60 & 80.00 \\
      %DNW & 76.00 & 80.00 \\
      %SNFS $+$ ERK & 75.20 & 80.00 \\
      %RigL $+$ ERK & 75.10 & 80.00 \\
      %STR & \textbf{76.19} & 79.55 \\
      %STR & 76.12 & \textbf{81.27} \\
      spred\footnotemark & $2e-5$ & - & $80.00$\\
      spred & $3e-6$ & - & $79.03$\\
      PILoT & $7e-6$ & $60000$ & $80.00$\\
      \hline
      %GMP & 73.91 & 90.00 \\
      %DNW & 74.00 & 90.00 \\
      %SNFS $+$ ERK & 72.90 & 90.00 \\
      %RigL $+$ ERK & 73.00 & 90.00 \\
      %STR & \textbf{74.73} & 87.7 \\
      %STR & 74.01 & 90.55 \\
      spred & $5e-6$ & - & $89.26$\\
      PILoT & $1e-5$ & $60000$ & $88.00$ \\
      PILoT & $1.4e-5$ & $60000$ & $91.00$\\
    \end{tabular}
    \end{minipage}
    \begin{minipage}{0.5\textwidth}
            \centering
    \begin{tabular}{l c c r } % <-- Alignments: 1st column left, 2nd middle and 3rd right, with vertical lines in between
      \textbf{Method} & $\alpha_{init}$ & $K$ & sparsity\\
      \hline
      %GMP & 70.59 & 95.00 \\
      %DNW & 68.30 & 95.00 \\
      %RigL $+$ ERK & 70.00 & 95.00 \\
      %STR & 70.4 & 95.03 \\
      spred & $2e-5$ & - & $94.50$\\
      PILoT & $2e-5$ & $60000$ & $94.00$\\
      PILoT & $3e-5$ & $60000$ & $95.00$\\
      PILoT & $3e-5$ & $60000$ & $95.60$ \\
      PILoT & $3e-5$ & $60000$ & $96.00$\\
      \hline
      %RigL $+$ ERK & 67.20 & 96.50 \\
      %STR & 67.22 & 96.53 \\
      spred & $3e-5$ & -& $97.19$\\
      PILoT & $4e-5$ & $40000$ & $97.19$ \\
      \hline
      %GMP & 57.90 & 98.00 \\
      %DNW & 58.20 & 98.00 \\
      %STR & 61.46 & 98.05 \\
      spred & $5e-5$ & - & $98.20$\\
      PILoT & $5e-5$ & $20000$ & 97.75 \\
      PILoT & $7e-5$& $20000$  &98.20\\
    \end{tabular}
    \end{minipage}
  %\end{center}
\end{table}
\footnotetext{Starting from a pretrained model with $77\%$ validation accuracy}
\paragraph{Label smoothing}
Altough PILoT is competitive in the $80\%-90\%$ sparsity range it is not SOTA.
Nevertheless, if we turn of label smoothing in the experiment PILoT outperforms STR in this range as well.
The only change for STR is turning of labelsmoothing.
For PILoT we use two different configurations.
We use $L_2$ regularization i.e. $||m \odot w||^2_{L_2}$ regularization set to $5 \cdot 10^{-5}$ instead of the value from STR experiment.
We use $K = 600000$ and $\delta =1.01$.
Furthermore, the strength of the PILoT regularization is initialized at $\{1 \cdot 10^{-5}, 2 \cdot 10^{-5}\}$ and no weight decay is used on the rest of the parameters.
The results are given in Table \ref{tab : res50 extra}.
\begin{table}[ht]
  \begin{center}
    \caption{Extra experiment ResNet-50 on ImageNet sparsity $(\%)$ versus accuracy $(\%)$ without label smoothing.}
    \label{tab:table1}
    \setlength{\tabcolsep}{2pt}
    \begin{tabular}{l c r } % <-- Alignments: 1st column left, 2nd middle and 3rd right, with vertical lines in between
      \textbf{Method} & \textbf{Top-1 Acc} & \textbf{Sparsity}\\
      ResNet-50 & 75.80 & 0\\
      \hline
      STR & 73.03 & \textbf{79.03} \\
      PILoT & \textbf{74.72} & \textbf{79.03} \\
      \hline
      STR & 71.6 & 89.26 \\
      PILoT & \textbf{73.21} & \textbf{91.41} \\
    \end{tabular}
  \end{center}
\end{table}\label{tab : res50 extra}

\subsection{Itterative Pruning}\label{appendix : itter}
\begin{figure}[H]   \centering
    \includegraphics[width=\textwidth]{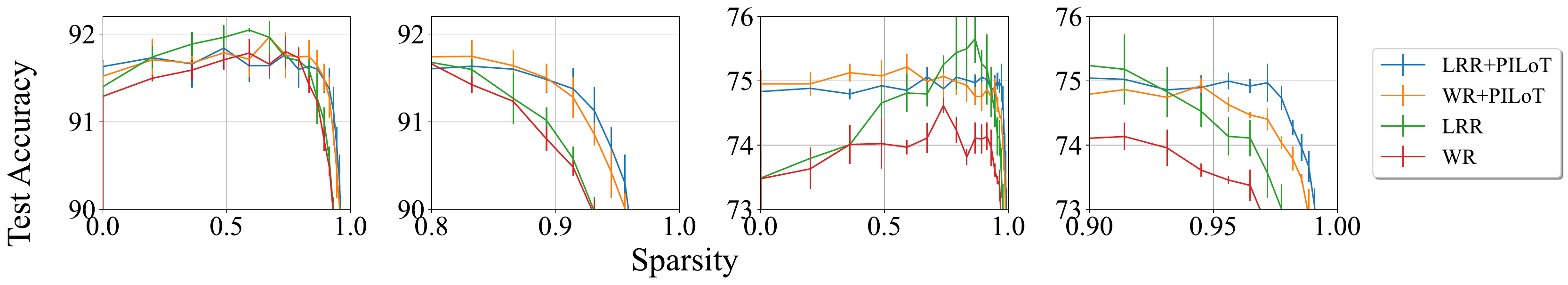}
    \caption{Learning Rate Rewinding (LRR) and Weight Rewinding (WR) with PILoT shows improvement over the baseline itterative pruning methods for CIFAR 10 and 100.}
    \label{fig: LRR CF}
\end{figure}

In Table \ref{experiment : IMP LRR im} the details of the ImageNet the experiment are given. 
Note the base learning rate $0.1$ is for the baseline and $0.2$ is used for our parameterization combined with scaling $1$. 
In addition, the $L_2$ regularization denotes the reparameterization of the original weight decay.
Thus for PILoT, in this case, we use that instead.
Moreover, all runs have been done for $3$ different seeds.
Furthermore, we provide additional experiments on CIFAR 10 and 100 with ResNet-20 and ResNet-18 respectively in Figure \ref{fig: LRR CF}.
The details are given in Table \ref{experiment : IMP LRR}

\begin{table}[ht]
\setlength{\abovecaptionskip}{5pt}
  \caption{WR and LRR experiment on ImageNet}
  \label{experiment : IMP LRR im}
  \centering
  \begin{tabular}{lll}
    \toprule
    %\multicolumn{2}{c}{Part}                   \\
    \cmidrule(r){1-2}
    Parameter     & Setting   & Comments  \\
    \midrule
    Optimizer &  SGD   &  \\
    Momentum & 0.9 & \\
    Batch size & 512 & \\
    Activation function & ReLu & \\
    Weight decay & $\{0,10^{-4}\}$ & \\
    Learning rate schedule    & step warmup  \\
    Base learning rate & $\{0.1, 0.2 \}$ &  \\ 
    Cycles     & $25$ &       \\
    Pruning rate & 0.8 & \\
    Epochs per cycle & $90$ & \\
    Warmup period & $10$ & \\
    Initialization & Kaiming normal & \\
    $L_2$ regularization & $5 \cdot 10^{-5}$ & Only for $m \odot w$ \\
    PILOT regularization & $\{0\}$ & Only for $m \odot w$\\
    Scaling & $1$ & Only for $m \odot w$ \\
    $\delta$ & 1 & Only for $m \odot w$ \\
    $K$ & $-$ &  Only for $m \odot w$ \\
    \bottomrule
    \end{tabular}
\end{table}

\begin{table}[ht]
\setlength{\abovecaptionskip}{5pt}
  \caption{WR and LRR experiment on CIFAR 10 and 100}
  \label{experiment : IMP LRR}
  \centering
  \begin{tabular}{lll}
    \toprule
    %\multicolumn{2}{c}{Part}                   \\
    \cmidrule(r){1-2}
    Parameter     & Setting   & Comments  \\
    \midrule
    Optimizer &  SGD   &  \\
    Momentum & 0.9 & \\
    Batch size & 256 & \\
    Activation function & ReLu & \\
    Weight decay & $10^{-4}$ & \\
    Learning rate schedule    & step warmup  \\
    Base learning rate & $\{0.1, 0.2 \}$ &  \\ 
    Cycles     & $25$ &       \\
    Pruning rate & 0.8 & \\
    Epochs per cycle & $150$ & \\
    Warmup period & $50$ & \\
    Initialization & Kaiming normal & \\
    $L_2$ regularization & $0$ & Only for $m \odot w$ \\
    PILOT regularization & $\{10^{-4}\}$ & Only for $m \odot w$\\
    Scaling & $1$ & Only for $m \odot w$ \\
    $\delta$ & 1 & Only for $m \odot w$ \\
    $K$ & $-$ &  Only for $m \odot w$ \\
    \bottomrule
    \end{tabular}
\end{table}

\section{Remark on Neuronwise Pruning}\label{section : neuronwise}
In the main text, we have used mirror flow to describe the implicit bias of parameterwise pruning. 
In this section, we show that neuronwise pruning can not be analyzed in the same way. 
We first define neuronwise pruning. 
Next, we paraphrase the necessary condition such that the implicit bias can be described by a mirror flow from \citep{Li2022ImplicitBO}. 
Finally, we show neuronwise pruning violates this condition pointing out a limitation of the framework. 

Consider the parameterization for a function with $p$ neurons, $g : \mathbb{R}^p \times \mathbb{R}^{n_1} \times \hdots \times \mathbb{R}^{n_p}$,
\begin{equation*}\label{ParamNeuronwise}
    g(m, w_1, \hdots, w_p) = \begin{pmatrix}
        m_1 w_1, \hdots, m_p w_p
    \end{pmatrix}
\end{equation*}
where $m_i \in \mathbb{R}$ is the mask and $w_i \in \mathbb{R}^{n_i}$ are the neurons. 

We state the necessary condition for a parameterization to induce a mirror flow.

\begin{theorem}\label{NessecaryConditionMflow} (Theorem 4.10 \citep{Li2022ImplicitBO})
    The Lie bracket span of $\{ \nabla_i g \}_{i = 1}^n$ is in the kernel of Jacobian $\partial g$.
\end{theorem}

Now, we use this theorem to show that neuronwise pruning does not induce a mirror flow. 

\begin{lemma}
    Neuronwise pruning violates Theorem \ref{NessecaryConditionMflow}.
\end{lemma}
Proof.
We show that for $p = 1$ the condition is already violated. This implies that in the general case, the condition is also violated as the neurons themselves are commuting with each other as they are parameterized separate.
 
To see this we can explicitly check the commuting condition for the following parameterization $g : \mathbb{R} \times \mathbb{R}^{n} \rightarrow \mathbb{R}^{n}$
\begin{equation*}
    g(m, w) = m w\end{equation*}
Then the gradients (Jacobians) and Hessian's are given by:
\begin{equation*}
     \nabla g_i = \begin{pmatrix}
         w_i \\
        m \mathbb{I}_{i=1}\\ 
        \vdots \\
            m \mathbb{I}_{i=n_1}
        \end{pmatrix} \qquad \text{and} \qquad Hg_i = \begin{pmatrix}
            0 & \mathbb{I}_{i=1} & \hdots & \mathbb{I}_{i=n} \\
             \mathbb{I}_{i=1} & 0 & \hdots & 0 \\
             \vdots & & & \vdots \\
              \mathbb{I}_{i=n} & 0 & \hdots & 0
        \end{pmatrix}
        \qquad \text{for } i = 1,2
    \end{equation*}
    Computing $ Hg_i \nabla g_j$
    \begin{equation*}
         Hg_i   \nabla g_j  = w_j \begin{pmatrix}
             0 \\
             \mathbb{I}_{i=1} \\
              \vdots \\
             \mathbb{I}_{i=n}
         \end{pmatrix}
    \end{equation*}
    we compute the Lie brackets which span a subspace of the Lie Algebra $LIE^{\geq 2}(\partial g)$. The subspace is spanned by
    \begin{equation*}
        \text{span}\left(w_j \begin{pmatrix}
             0 \\
             \mathbb{I}_{i=1} \\
              \vdots \\
             \mathbb{I}_{i=n}
         \end{pmatrix}- w_i \begin{pmatrix}
             0 \\
             \mathbb{I}_{j=1} \\
              \vdots \\
             \mathbb{I}_{j=n}
         \end{pmatrix} \text{ for } i , j \in [n]\right) \subset  LIE^{\geq 2}(\partial g).
    \end{equation*}
    Clearly, this span is not in $Ker(\partial g)$ as can be shown by a direct computation:
    \begin{equation*}
        \left(\partial g \right) \left(
        w_j \begin{pmatrix}
             0 \\
             \mathbb{I}_{i=1} \\
              \vdots \\
             \mathbb{I}_{i=n}
         \end{pmatrix}- w_i \begin{pmatrix}
             0 \\
             \mathbb{I}_{j=1} \\
              \vdots \\
             \mathbb{I}_{j=n}
         \end{pmatrix} \right) = 
        \begin{pmatrix}
            \mathbb{I}_{i=1} m w_j - \mathbb{I}_{j=1} m w_i, \hdots,  \mathbb{I}_{i=n}m w_i - \mathbb{I}_{j=1} m w_i
        \end{pmatrix} 
    \end{equation*}
    For the span to be in the kernel we need either that all $w_i = 0$ for all $i \in [n]$ or $m = 0$. In both cases this implies that $g(m, w) \in \{ 0 \}$. This implies that a mirror flow is only well-defined if the parameterization is zero. Hence, neuron-wise continuous sparsification does not induce a mirror flow. $\square$

\end{document}